\documentclass[11pt,a4paper]{article}
\usepackage[hyperref]{emnlp2020}
\usepackage{times}
\usepackage{latexsym}

\usepackage{microtype}

\aclfinalcopy %

\usepackage{color,xcolor}
\usepackage{epsfig}
\usepackage{graphicx}

\usepackage{adjustbox}
\usepackage{array}
\usepackage{booktabs}
\usepackage{colortbl}
\usepackage{float,wrapfig}
\usepackage{hhline}
\usepackage{multirow}
\usepackage{subcaption} %

\usepackage{amsmath,amsfonts,amsthm,amssymb}
\usepackage{bm}
\usepackage{nicefrac}
\usepackage{microtype}

\usepackage{changepage}
\usepackage{extramarks}
\usepackage{fancyhdr}
\usepackage{lastpage}
\usepackage{setspace}
\usepackage{soul}
\usepackage{xspace}

\usepackage{algorithm, algorithmic}
\usepackage{enumerate}

\usepackage{titlesec}

\usepackage{mdframed}
\usepackage{listings}
\usepackage{verbatim}
\usepackage{enumitem}
\setlist[enumerate]{itemsep=0mm}

\newcolumntype{L}[1]{>{\raggedright\let\newline\\\arraybackslash\hspace{0pt}}m{#1}}
\newcolumntype{C}[1]{>{\centering\let\newline\\\arraybackslash\hspace{0pt}}m{#1}}
\newcolumntype{R}[1]{>{\raggedleft\let\newline\\\arraybackslash\hspace{0pt}}m{#1}}

\newcommand{\sect}[1]{Section~\ref{#1}}
\newcommand{\ssect}[1]{\S~\ref{#1}}

\newcommand{\fig}[1]{Figure~\ref{#1}}
\newcommand{\tbl}[1]{Table~\ref{#1}}

\newcommand{\ignorethis}[1]{}

\newcommand{\ignore}[1]{}

\makeatletter
\DeclareRobustCommand\onedot{\futurelet\@let@token\@onedot}
\def\@onedot{\ifx\@let@token.\else.\null\fi\xspace}

\makeatother

\definecolor{MyDarkBlue}{rgb}{0,0.08,1}
\definecolor{MyDarkGreen}{rgb}{0.02,0.6,0.02}
\definecolor{MyDarkRed}{rgb}{0.8,0.02,0.02}
\definecolor{MyDarkOrange}{rgb}{0.40,0.2,0.02}
\definecolor{MyDarkYellow}{rgb}{0.40,0.4,0.02}
\definecolor{MyPurple}{RGB}{111,0,255}
\definecolor{MyRed}{rgb}{1.0,0.0,0.0}
\definecolor{MyGold}{rgb}{0.75,0.6,0.12}
\definecolor{MyDarkgray}{rgb}{0.66, 0.66, 0.66}

\newcommand{\model}{CALM\xspace}
\newcommand{\fullmodel}{Contextual Action Language Model\xspace}

\title{Keep CALM and Explore:\\Language Models for Action Generation in Text-based Games}

\author{{Shunyu Yao$^\dagger$, Rohan Rao$^\dagger$, Matthew Hausknecht$^\ddagger$, Karthik Narasimhan$^\dagger$} \\
$^\dagger$Princeton University \quad\quad  $^\ddagger$ Microsoft Research \\
\texttt{\{shunyuy, rohanr, karthikn\}@princeton.edu} \\
\texttt{matthew.hausknecht@microsoft.com} \\
}

\date{}

\begin{document}
\maketitle

\begin{abstract}

Text-based games present a unique challenge for autonomous agents to operate in natural language and handle enormous action spaces. In this paper, we propose the \fullmodel (\model) to generate a compact set of action candidates at each game state. Our key insight is to train language models on human gameplay, where people demonstrate linguistic priors and a general \emph{game sense} for promising actions conditioned on game history. We combine \model with a reinforcement learning agent which re-ranks the generated action candidates to maximize in-game rewards. 
We evaluate our approach using the Jericho benchmark~\cite{hausknecht19colossal}, on games \emph{unseen} by \model during training. Our method obtains a 69\% relative improvement in average game score over the previous state-of-the-art model. Surprisingly, on half of these games, \model is competitive with or better than other models that have access to ground truth admissible actions.\footnote{Code and data are available at \url{https://github.com/princeton-nlp/calm-textgame}.}

\end{abstract}

\section{Introduction}

\begin{figure}[ht]
\begin{mdframed}
  \parbox{\textwidth}{
\emph{Observation:} 
You are in the living room. There is a doorway to the east, a wooden door with strange gothic lettering to the west, which appears to be nailed shut, a trophy case, and a large oriental rug in the center of the room. You are carrying: A brass lantern \dots
}
\begin{mdframed}

\begin{flushleft}

\textbf{Random Actions}: 

{\color{darkgray}close door, north a, eat troll with egg, \dots} 

\textbf{\model (n-gram) Actions}:

{\color{darkgray}enter room, leave room, lock room,}

{\color{darkgray}open door, close door, knock on door, \dots}

\textbf{\model (GPT-2) Actions}: 

{\color{darkgray} east, open case, get rug, turn on lantern, \underline{\color{black} move rug}, unlock case with key,  \dots}

\end{flushleft}
\end{mdframed}

\emph{Next Observation}: With a great effort, the rug is moved to one side of the room, revealing the dusty cover of a closed trap door...
  
\end{mdframed}
\caption{Sample gameplay from \textsc{Zork1} along with action sets generated by two variants of \model. The game recognizes a vocabulary size of 697, resulting in more than $697^4 \approx$ 200 billion potential 4-word actions. `\emph{move rug}' is the optimal action to take here and is generated by our method as a candidate.
}
    \label{fig:teaser}
     \vspace{-6mm}
 \end{figure}

Text-based games have proven to be useful testbeds for developing agents that operate in language. As interactions in these games (input observations, action commands) are through text, they require solid language understanding for successful gameplay. While several reinforcement learning (RL) models have been proposed recently~\cite{narasimhan15,he2015deep,hausknecht19colossal,ammanabrolu2019playing}, combinatorially large action spaces continue to make these games challenging for these approaches.

The action space problem is exacerbated by the fact that only a tiny fraction of action commands are admissible in any given game state. An admissible action is one that is parseable by the game engine and changes the underlying game state. For example, in Figure~\ref{fig:teaser}, one can observe that randomly sampling actions from the game vocabulary leads to several inadmissible ones like `\emph{north a}'  or  `\emph{eat troll with egg}’. Thus, narrowing down the action space to admissible actions requires both syntactic and semantic knowledge, making it challenging for current systems. 

Further, even within the space of admissible actions, it is imperative for an autonomous agent to know which actions are most promising to advance the game forward, and explore them first. Human players innately display such game-related common sense. For instance in Figure~\ref{fig:teaser}, players might prefer the command ``move rug'' over ``knock on door''  since the door is nailed shut. However, even the state-of-the-art game-playing agents do not incorporate such priors, and instead rely on rule-based heuristics~\cite{hausknecht19colossal} or handicaps provided by the learning environment~\cite{hausknecht19colossal,ammanabrolu2020graph} to circumvent these issues.

In this work, we propose the \fullmodel (\model) to alleviate this challenge. Specifically, at each game step we use \model to generate action candidates, which are fed into a Deep Reinforcement Relevance Network (DRRN)~\cite{he2015deep} that uses game rewards to learn a value function over these actions. This allows our model to combine generic linguistic priors for action generation with the ability to adaptively choose actions that are best suited for the game.

To train \model, we introduce a novel dataset of 426 human gameplay transcripts for 590 different text-based games. While these transcripts are noisy and actions are not always optimal, they contain a substantial amount of linguistic priors and game sense. Using this dataset, we train a single instance of \model and deploy it to generate actions across many different downstream games. Importantly, in order to demonstrate the generalization of our approach, we do not use any transcripts from our evaluation games to train the language model.

We investigate both n-gram and state-of-the-art GPT-2~\cite{radford2019language} language models and first evaluate the quality of generated actions in isolation by comparing against ground-truth sets of admissible actions. Subsequently, we evaluate the quality of \model in conjunction with RL over 28 games from the Jericho benchmark~\cite{hausknecht19colossal}. Our method outperforms the previous state-of-the-art method by 69\% in terms of average normalized score. Surprisingly, on 8 games our method even outperforms competing methods that use the admissible action handicap -- for example, in the game of \textsc{inhumane}, we achieve a score of 25.7 while the state-of-the-art KG-A2C agent~\citep{ammanabrolu2020graph} achieved 3.

In summary, our contributions are two-fold. First, we propose a novel learning-based approach for reducing enormous action spaces in text-based games using linguistic knowledge. Second, we introduce a new dataset of human gameplay transcripts, along with an evaluation scheme to measure the quality of action generation in these games.

\section{Related Work}
\begin{figure*}[t]
    \centering
    \includegraphics[width=\textwidth]{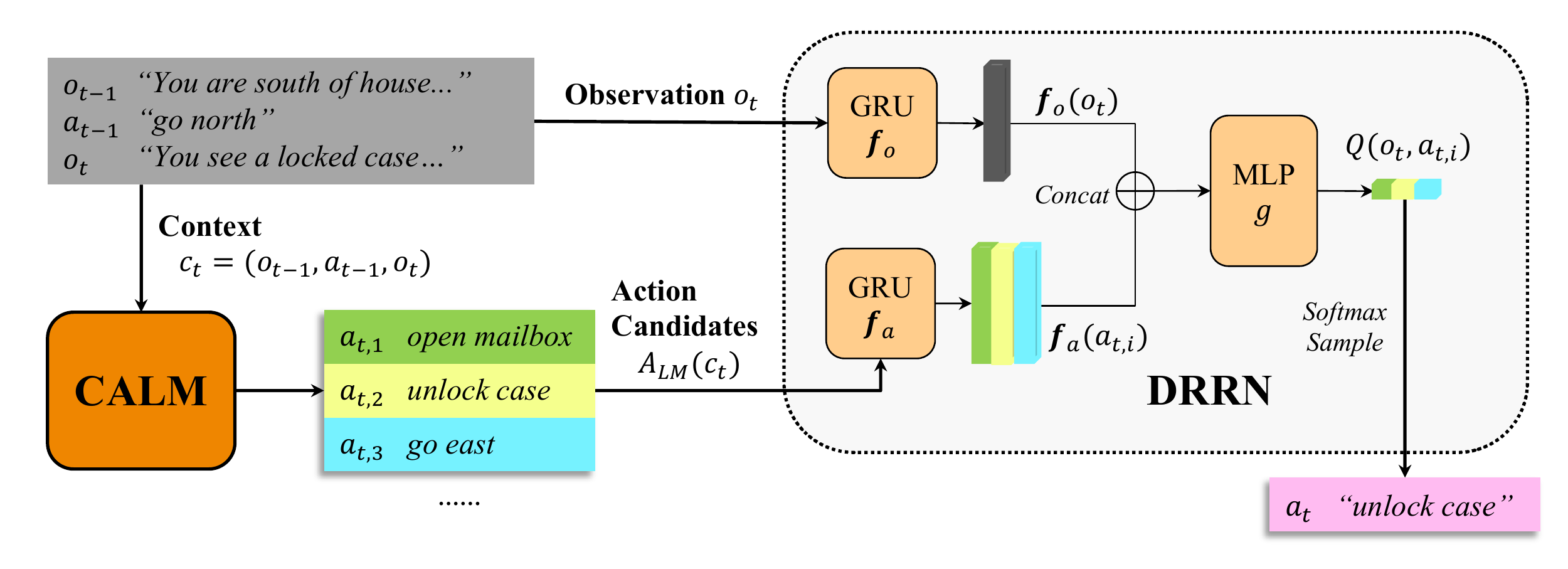}
    \caption{\model combined with an RL agent -- DRRN~\cite{he2015deep}  -- for gameplay. \model is trained on transcripts of human gameplay for action generation. At each state, \model generates action candidates conditioned on the game context, and the DRRN calculates the Q-values over them to select an action. Once trained, a single instance of \model can be used to generate actions for any text-based game.}
    \label{fig:framework}
\end{figure*}

\paragraph{Reinforcement Learning for Text-based Games} Early work on text-based games~\cite{narasimhan15,he2015deep} developed RL agents on synthetic environments with small, pre-defined text action spaces. Even with small actions spaces (e.g. $<200$ actions), approaches to filter inadmissible actions~\cite{zahavy18,jain2019} led to faster learning convergence.
Recently, \citet{hausknecht19colossal} introduced Jericho -- a benchmark of challenging man-made text games. These games contain significantly greater linguistic variation and larger action spaces compared to frameworks like TextWorld~\cite{cote2018textworld}. 

To assist RL agents, Jericho provides a handicap that identifies admissible actions at each game state. This has been used by approaches like DRRN~\cite{he2015deep} as a reduced action space. Other RL agents like TDQN~\cite{hausknecht19colossal} and KGA2C~\cite{ammanabrolu2020graph} rely on the handicap for an auxiliary training loss.
In general, as these RL approaches lack linguistic priors and only learn through in-game rewards, they are reliant on the admissible-action handicap to make the action space tractable to explore. %

\paragraph{Linguistic Priors for Text-based Games} A different line of work has explored various linguistic priors for generating action commands. \citet{fulda17} used Word2vec~\cite{mikolov13} embeddings to infer affordance properties (i.e.\,verbs suitable for an object). 
Other approaches~\cite{kostka17, hausknecht2019nail} trained simple n-gram language models to learn affordances for action generation.
Perhaps most similar to our work is that of \citet{tao2018towards}, who trained seq2seq~\cite{sutskever2014sequence} models to produce admissible actions in synthetic TextWorld~\cite{cote2018textworld} games. 
In a slightly different setting, \citet{urbanek2019light} trained BERT~\cite{devlin2018bert} to generate contextually relevant dialogue utterances and actions in fantasy settings. 
However, these approaches are game-specific and do not use any reinforcement learning to optimize gameplay. In contrast, we combine strong linguistic priors with reinforcement learning, and use a modern language model that can generate complex actions and flexibly model the dependency between actions and contexts. We also train on multiple games and generalize to unseen games.

\paragraph{Generation in Text-based Games and Interactive Dialog} Besides solving games, researchers have also used language models to \emph{create} text-based games. \citet{ammanabrolu2019toward} used Markov chains and neural language models to procedurally generate quests for TextWorld-like games. AI Dungeon 2~\cite{walton} used GPT-2 to generate narrative text in response to arbitrary text actions, but lacked temporal consistency over many steps.

More broadly, the concept of generating candidates and re-ranking has been studied in other interactive lanugage tasks such as dialogue~\cite{zhao-eskenazi-2016-towards,williams-etal-2017-hybrid,song2016two,chen2017survey} and communication games~\cite{lazaridou2020multi}. 
These approaches often focus on improving aspects like fluency and accuracy of the generated utterances, whereas our re-ranking approach only aims to maximize future rewards in the task. Also, 
our \model pre-trained model generalizes to new environments without requiring any re-training.

\section{Method}

\subsection{Background}
\label{background}
A text-based game can be formally specified as a partially observable Markov decision process (POMDP) $(S, T, A, O, R, \gamma)$, where a player issues text actions $a \in A$ and receives text observations $o \in O$ and scalar rewards $r = R(s, a)$ at each step.
Different games have different reward designs, but typically provide sparse positive rewards for solving key puzzles and advancing the story, and negative rewards for dying. $\gamma \in [0,1]$ is the reward discount factor. Latent state $s \in S$ contains the current game information (e.g.\,locations of the player and items, the player's inventory), which is only partially reflected in $o$. The transition function $s'=T(s, a)$ specifies how action $a$ is applied on state $s$, and $a$ is \textbf{admissible} at state $s$ if $T(s, a) \neq s$ (i.e.\,if it is parseable by the game and changes the state). $S$, $T$ and $R$ are not provided to the player.

\paragraph{Reinforcement Learning}
One approach to developing text-based game agents is reinforcement learning (RL). The Deep Reinforcement Relevance Network (DRRN)~\cite{he2015deep} is an RL algorithm that learns a Q-network $Q_\phi(o, a)$ parametrized by $\phi$. The model encodes the observation $o$ and each action candidate $a$ using two separate encoders $f_o$ and $f_a$ (usually recurrent neural networks such as GRU \cite{cho-etal-2014-properties}), and then aggregates the representations to derive the Q-value through a decoder $g$:
\begin{equation}
    Q_\phi(o, a) = g(f_o(o), f_a(a))
    \label{drrn}
\end{equation}
For learning $\phi$, tuples $(o, a, r, o')$ of observation, action, reward and the next observation are sampled from an experience replay buffer and the following temporal difference (TD) loss is minimized:
\begin{equation}
   \mathcal{L}_\text{\scalebox{.8}{TD}}(\phi) = (r + \gamma \max_{a' \in A} Q_\phi(o', a') - Q_\phi(o, a))^2
   \label{td}
\end{equation}
During gameplay, a softmax exploration policy is used to sample an action:
\begin{equation}
    \pi_\phi(a|o) = \frac{\exp(Q_\phi(o, a))}{\sum_{a' \in A} \exp(Q_\phi(o, a'))}
    \label{softmax}
\end{equation}
While the above equation contains only a single observation, this can also be extended to a policy $\pi(a|c)$ conditioned on a longer context $c = (o_1, a_1, ... , o_t) $ of previous observations and actions till current time step $t$. Note that when the action space $A$ is large, \eqref{td} and \eqref{softmax} become intractable.

\subsection{\fullmodel (\model)}
To reduce large action spaces and make learning tractable, we train language models to generate compact sets of actions candidates. Consider a dataset $\mathcal{D}$ of $N$ trajectories of human gameplay across different games, where each trajectory of length $l$ consists of interleaved observations and actions $(o_1, a_1, o_2, a_2, \cdots, o_{l}, a_{l})$. 
The \emph{context} $c_t$ at timestep $t$ is defined as the history of observations and actions, i.e. $c_t = (o_1, a_1, ... , a_{t-1}, o_t)$. In practice, we find that a window size of 2 works well, i.e. $c_t = (o_{t-1}, a_{t-1}, o_t)$. We train parametrized language models $p_\theta$ to generate actions $a$ conditioned on contexts $c$. Specifically, we use all $N$ trajectories and minimize the following cross-entropy loss:
\begin{equation}
\label{eq:train0}
    \mathcal{L}_\text{\scalebox{.8}{LM}}(\theta) = - \mathbb{E}_{(a, c) \sim D}\log p_\theta(a | c)
\end{equation}
Since each action $a$ is typically a multi-word phrase consisting of $m$ tokens $a^1, a^2, \cdots, a^m$, we can further factorize the right hand side of \eqref{eq:train0} as:
\begin{equation}
\label{eq:train1}
    p_\theta(a|c) = \prod_{i=1}^m p_\theta(a^i | a^{<i}, c)
\end{equation}
Thus, we can simply use the cross-entropy loss over each token $a^i$ in action $a$ during training. We investigate two types of language models: 

\textbf{1. n-gram}: This model simply uses n-gram counts from actions in $\mathcal{D}$ to model the following probability:
\begin{equation}
    p_{(n,\alpha)}(a^i | a^{<i}) = 
     \frac{cnt(a^{i-n+1}, \cdots, a^{i}) + \alpha}{cnt(a^{i-n+1}, \cdots, a^{i-1}) + \alpha |V|}
\end{equation} 
where $cnt(a^i, \cdots, a^{j})$ counts the number of occurrences of the action sub-sequence $(a^i, \cdots, a^{j})$ in the training set, $\alpha$ is a smoothing constant, and $V$ is the token vocabulary. 
Note that this model is trained in a \emph{context-independent} way and only captures basic linguistic structure and common affordance relations observed in human actions. We optimize the parameters $(n, \alpha)$ to minimize the perplexity on a held-out validation set of actions. 

To generate top actions given context $c$, we construct a restricted action space $\mathcal{A}_c = \mathcal{V} \times \mathcal{B}_c$, where $\mathcal{V}$ is the set of verb phrases (e.g.\,\emph{open}, \emph{knock on}) collected from training actions, and $\mathcal{B}_c$ is the set of nouns (e.g.\,\emph{door}) detected in $c$ using spaCy's\footnote{\url{https://spacy.io/}} noun-phrase detection. Then we calculate $p_{(n,\alpha)}(a)$ for each $a \in \mathcal{A}_c$ and choose the top ones.

\textbf{2. GPT-2}~\cite{radford2019language}: We use a pretrained GPT-2 and train it on $\mathcal{D}$ according to \eqref{eq:train0} and \eqref{eq:train1}.
Unlike the previous n-gram model, GPT-2 helps model dependencies between the context and the action in a flexible way, relying on minimal assumptions about the structure of actions. We use beam search to generate most likely actions.

\subsection{Reinforcement Learning with CALM}
\label{rl}
Though language models learn to generate useful actions, they are not optimized for gameplay performance.
Therefore, we use \model to generate top-$k$ action candidates $A_\text{{LM}}(c, k) \subset A$ given context $c$, and train a DRRN to learn a Q-function over this action space. This can be done by simply replacing  $A$ with $A_\text{{LM}}(c, k)$ in equations  \eqref{td} and \eqref{softmax}. In this way, we combine the \model's generic action priors with the ability of RL to learn policies optimized for the gameplay.
We choose not to fine-tune \model in RL so as to avoid overfitting to a specific game and invalidate the general priors present in \model.

To summarize, we employ \model for providing a reduced action space for text adventure agents to explore efficiently. Even though we choose a specific RL agent (DRRN) in our experiments, \model is simple and generic, and can be combined with any RL agent.

\section{Experimental Setup}

We perform empirical studies to 1) evaluate the quality of actions generated by \model in isolation from the complexities of RL, 2) evaluate \model combined with an RL agent for gameplay, and 3) analyze what factors contribute to the effectiveness of our method. We describe our setup in this section and provide results in \sect{sec:results}.

\subsection{Data and Environment}
\begin{figure}[t]
\centering
\includegraphics[width=.49\textwidth]{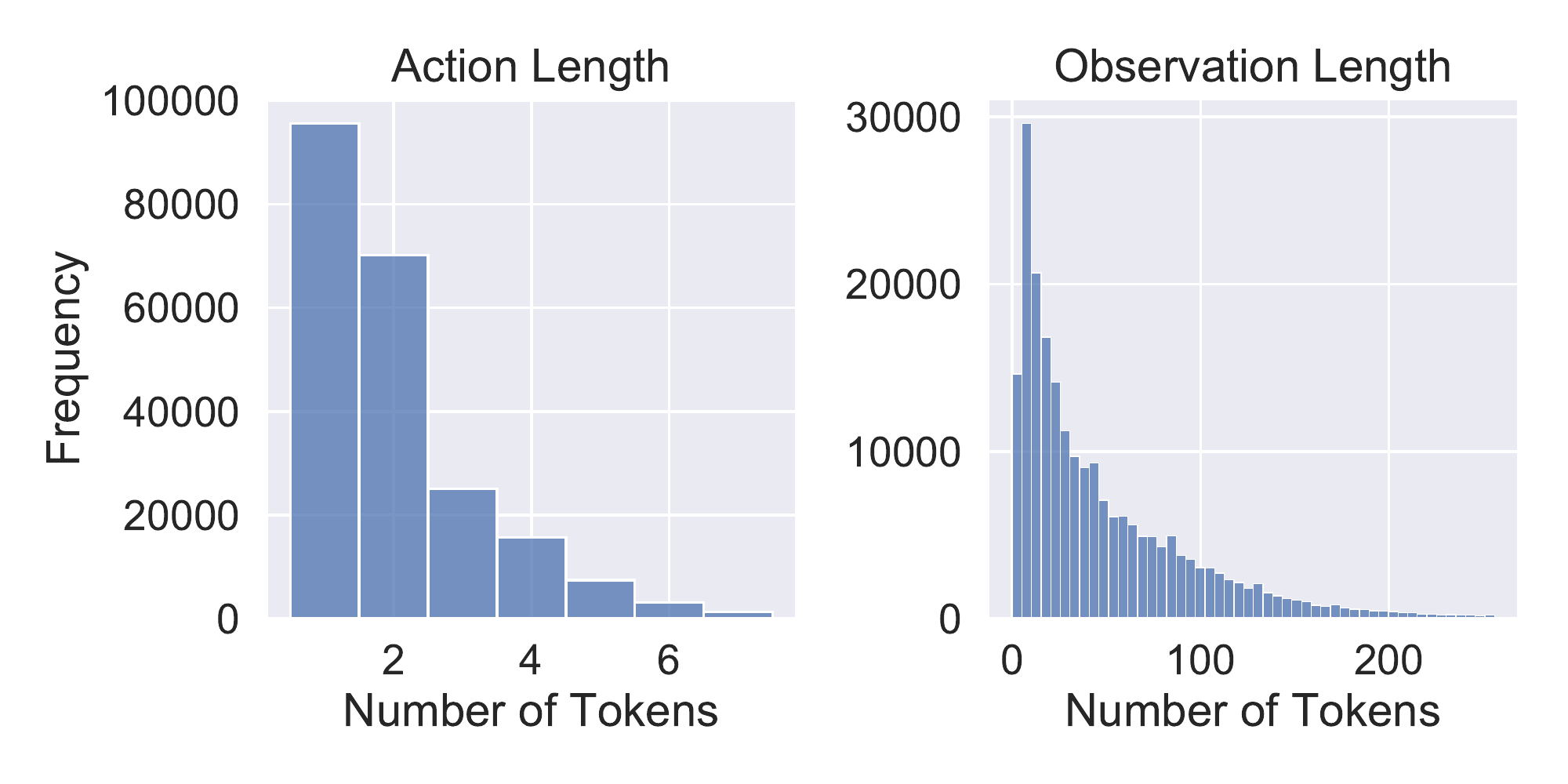}
\caption{Distributions of actions and observations in the ClubFloyd Dataset, in terms of the number of tokens. Actions more than 7 tokens ($<$0.5\%) and observations more than 256 tokens ($<$2\%) are trimmed.}
\label{fig:dataset_profile}
\end{figure}

\paragraph{ClubFloyd Dataset} We collect data from ClubFloyd\footnote{\url{http://www.allthingsjacq.com/interactive_fiction.html\#clubfloyd}}, which archives transcripts of humans cooperatively playing text-based games. We crawl 426 transcripts covering 590 games (in some transcripts people play more than one game), and build a dataset of 223,527 context-action pairs  $\{((o_{t-1}, a_{t-1}, o_{t}), a_{t})\}$. We pre-process the data by removing samples with meta-actions (e.g.\,`\emph{save}','\emph{restore}') or observations with over 256 tokens. \fig{fig:dataset_profile} visualizes the action and observation length distributions. 
We also note that a few common actions (e.g.\,`\emph{north}', `\emph{take all}', `\emph{examine}') make up a large portion of the data. More details on the dataset are in the supplementary material.

\paragraph{Game Environment} 

To test our RL agents, we use 28 man-made text games from the Jericho framework~\cite{hausknecht19colossal}. We augment state observations with location and inventory descriptions by issuing the `\emph{look}' and `\emph{inventory}' commands, following the standard practice described in ~\citet{hausknecht19colossal}. 

The Jericho framework implements an \emph{admissible action handicap} by enumerating all combinations of game verbs and  objects at each state, and testing each action's admissibility by accessing the underlying simulator states and load-and-save functions. As a result, the handicap runs no faster than a GPT-2 inference pass, and could in fact be unavailable for games outside Jericho. Jericho also provides an optimal \emph{walkthrough} trajectory to win each game. \tbl{tab:ClubFloydData} provides statistics of the ClubFloyd Dataset and the Jericho walkthroughs. We observe that ClubFloyd has a much larger vocabulary and a diverse set of games, which makes it ideal for training \model.
We utilize Jericho walkthroughs in our standalone evaluation of \model in \ssect{wt}.

\subsection{\model Setup}

\paragraph{Training} 
For training \model (n-gram), we condition only on the current observation, i.e.\,$c_t=o_t$ instead of $c_t=(o_{t-1}, a_{t-1}, o_t)$, since $o_{t-1}$ and $a_{t-1}$ may contain irrelevant objects to the current state. We split the dataset into 90\% training set and 10\% validation set, and choose $n$ and $\alpha$ based on the validation set perplexity. We find a bi-gram model $n = 2, \alpha = 0.00073$ works best, achieving a per-action perplexity of $863,808$ on the validation set and $17,181$ on the training set. 

For \model (GPT-2), we start with a 12-layer, 768-hidden, 12-head, 117M parameter GPT-2 model pre-trained on the WebText corpus~\cite{radford2019language}. The implementation and pretrained weights of this model are obtained from \citet{Wolf2019HuggingFacesTS}. We then train it on the ClubFloyd transcripts for 3 epochs to minimize ~\eqref{eq:train0}. We split the dataset into 90\% training set and 10\% validation set and we obtain a training loss of 0.25 and a validation loss of 1.98. Importantly, \emph{both models are trained only on transcripts that do not overlap with the 28 Jericho games we evaluate on}.

\begin{figure*}[t]
    \centering
    \includegraphics[width=0.9\textwidth]{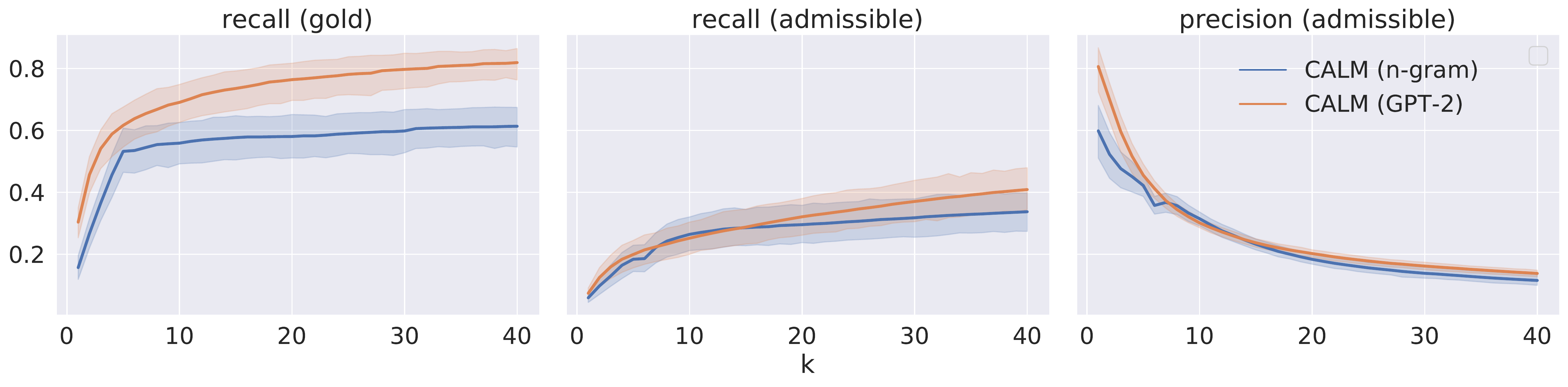}
    \caption{Precision and recall of gold and admissible actions generated by \model, evaluated on walkthrough trajectories of 28 games provided by Jericho. $k$ is the number of actions generated by \model. Shaded areas represent standard deviation.}
    \label{fig:wt}
\end{figure*}

\paragraph{Generating Top Actions}
For every unique state of each game, we generate the top $k=30$ actions. For \model (n-gram), we enumerate all actions in $\mathcal{A}_c$ plus 13 one-word directional actions (e.g.\,`\emph{north}', `\emph{up}', `\emph{exit}'). To encourage action diversity, at most 4 actions are generated for each object $b\in \mathcal{B}_c$. For \model (GPT-2), we use beam search with a beam size of $40$, and then choose the top $30$ actions.

\begin{table}[t]
    \centering
\resizebox{\columnwidth}{!}{
    \begin{tabular}{l|ll}
    \multirow{2}{*}{} & \textbf{ClubFloyd} & \textbf{Jericho} \\
    & \textbf{Dataset} & \textbf{Walkthroughs} \\ \midrule
        \textit{\# unique games} & 590 & 28 \\
        \textit{Vocab size} & 39,670 & 9,623  \\ 
        \textit{Vocab size (game avg.)} & 2,363 &1,037 \\
        \textit{Avg.\,trajectory length} & 360  & 98 \\
        \textit{Action Quality} & Non-optimal & Optimal \\
        \end{tabular}
        }
    \caption{Statistics of the ClubFloyd Dataset and Jericho walkthrough trajectories.}
        \label{tab:ClubFloydData}
\end{table}

\subsection{RL Agent Setup}
\paragraph{Training}
We use DRRN~\cite{he2015deep} to estimate Q-Values over action candidates generated by \model. Following \citet{hausknecht19colossal}, we use a FastText model~\cite{joulin2017bag} to predict the admissibility of an action based on the game's textual response and filter out candidate actions that are found to be inadmissible.
We train the DRRN asynchronously on 8 parallel instances of the game environment for $10^6$ steps in total. Following \citet{narasimhan15},  we use a separate experience replay buffer to store trajectories with the best score at any point of time. The final score of a training run is taken to be the average score of the final 100 episodes during training. For each game, we train five independent agents with different random seeds and report the average score. For model variants in \ssect{sec:analysis} we only run one trail.

\paragraph{Baselines} We compare with three baselines:

	1. \emph{NAIL}~\citep{hausknecht2019nail}: Uses hand-written rules to act and explore, therefore requires no reinforcement learning or oracle access to admissible actions.
	
 	2. \textit{DRRN}~\cite{he2015deep}: This RL agent described in \ssect{background} uses ground-truth admissible actions provided by the Jericho handicap.
 	
	3. \textit{KG-A2C}~\cite{ammanabrolu2020graph}: This RL agent constructs a game knowledge graph to augment the state space as well as constrain the types of actions generated. During learning, it requires the admissible action handicap to guide its exploration of the action space.
	
Of these methods, DRRN and KG-A2C require ground-truth admissible actions, which our model does not use, but we add them as reference comparisons for completeness.

\section{Results}
\label{sec:results}

\subsection{Evaluating \model on walkthroughs}
\label{wt}
Metrics like validation loss or accuracy on validation set of our ClubFloyd data are not sufficient to evaluate \model (see supplementary material for details on these metrics). This is because: 1) there can be multiple admissible actions in each state, and 2) the human actions in the trajectories are not guaranteed to be optimal or even admissible. Therefore, we use the walkthroughs provided in Jericho to provide an additional assessment on the quality of actions generated by \model.

Consider a walkthrough to be an optimal trajectory $(o_1, a_1, \cdots, o_l, a_l)$ leading to the maximum score achievable in the game. At step $t$ ($1 \le t \le l$), the context $c_t$ is $(o_{t-1}, a_{t-1}, o_t)$, the \emph{gold} action is $a_t$ and the full set of admissible actions $A_t$ is obtained from the Jericho handicap. 
Suppose the generated set of top-$k$ actions at step $t$ is $A_\text{{LM}}(c_t, k)$. We then calculate the average precision of admissible actions ($prec_a$), recall of admissible actions ($rec_a$), and recall of gold actions ($rec_g$) as follows:
\begin{align}
prec_a(k) &= \frac{1}{l} \sum_{t=1}^l{\frac{\left|A_t \cap A_\text{{LM}}(c_t, k)\right|}{k}} \\
rec_a(k) &= \frac{1}{l} \sum_{t=1}^l{\frac{\left|A_t \cap A_\text{{LM}}(c_t, k)\right|}{|A_t|}} \\
rec_g(k) &= \frac{1}{l} \sum_{t=1}^l{ |\{a_t\} \cap A_\text{{LM}}(c_t, k)|}
\label{eq:p_v}
\end{align}

We calculate these metrics on each of the 28 games and present the averaged metrics as a function of $k$ in \fig{fig:wt}. The $rec_a$ curve shows that the top $k=15$ actions of \model (GPT-2 and n-gram) are both expected to contain around 30\% of all admissible actions in each walkthrough state. However, when $k$ goes from 15 to 30, \model (GPT-2) can come up with 10\% more admissible actions, while the gains are limited for \model(n-gram). When $k$ is small, \model (n-gram) benefits from its strong action assumption of one verb plus one object. However, this assumption also restricts \model (n-gram) from generating more complex actions (e.g.\,`open case with key') that \model (GPT-2) can produce. This can also be seen in the $rec_g$ curve, where the top-30 actions from \model (GPT-2) contain the gold action in $20\%$ more game states than \model (n-gram). This gap is larger when it comes to gold actions, because they are more likely to be complex actions that the \model(n-gram) is unable to model. 

Further, we note that as $k$ increases, the average quality of the actions decreases ($prec_a$ curve), while they contain more admissible actions ($rec_a$ curve). Thus, $k$ plays an important role in balancing exploration (more admissible actions) with exploitation (a larger ratio of admissible actions) for the RL agent, which we demonstrate empirically in \ssect{sec:analysis}. We provide several examples of generated actions from both models in the supplementary material.

\subsection{Evaluating gameplay on Jericho}
\begin{figure*}[t]
    \centering
    \includegraphics[width=0.878\textwidth]{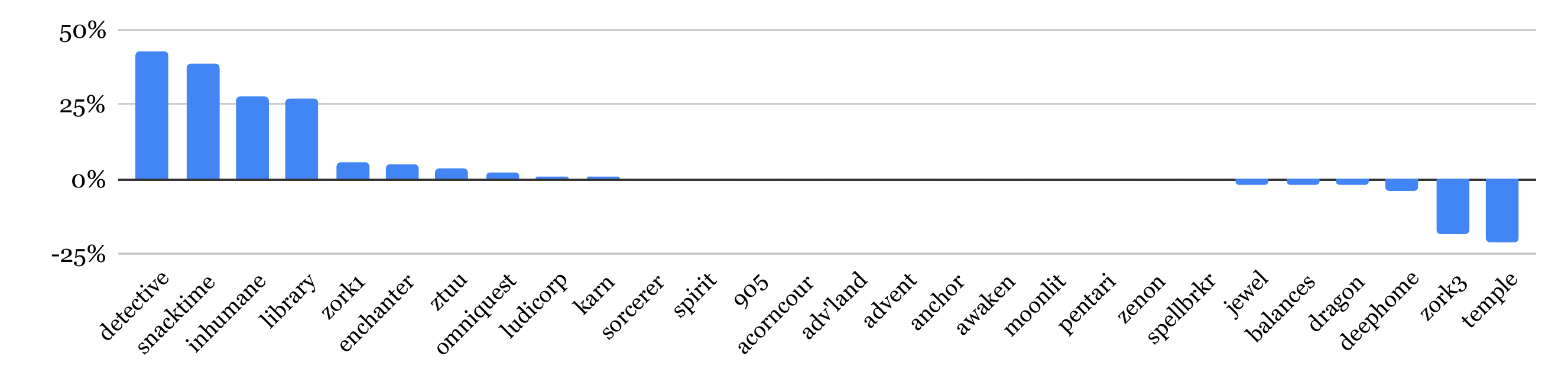}
    \caption{Difference in normalized scores achieved by \model(GPT-2) and NAIL, in decreasing order.}
    \label{fig:chart}
\end{figure*}
\begin{table}[t]
    \centering
\resizebox{\columnwidth}{!}{
    \begin{tabular}{r|lll|ll|l}
 & \multicolumn{3}{c|}{\textit{Without Handicap}}  & \multicolumn{2}{c|}{\textit{With Handicap}} & \\
 \midrule
\multirow{2}{*}{\textbf{Game}} & \textbf{\model} & \textbf{\model} & \textbf{NAIL} & \textbf{KG-A2C} & \textbf{DRRN} & \textbf{Max} \\
    & \textbf{(GPT-2)} & \textbf{(n-gram)} & & & & \textbf{Score} \\
\midrule
\midrule
905 & 0 & 0 & 0 & 0 & 0 & 1 \\
acorncourt   & 0 & 0 & 0 & 0.3 & 10 & 30 \\
advland   & 0 & 0 & 0 & 0 & 20.6 & 100 \\
advent & 36 & 36 & 36 & 36 & 36 & 350 \\
anchor   & 0 & 0 & 0 & 0 & 0 & 100 \\
awaken  & 0 & 0 & 0 & 0 & 0 & 50 \\
balances  & 9.1 & 8.9 & \textbf{10} & 10 & 10 & 51 \\
deephome   & 1.0 & 1.0 & \textbf{13.3} & 1.0 & 1.0 & 300 \\
\underline{detective}  & \textbf{289.7} & 284.3 & 136.9 & 207.9 & 197.8 & 360 \\
\underline{dragon} & 0.1 & 0.0 & \textbf{0.6} & 0 & -3.5 & 25 \\
enchanter  & \textbf{19.1} & 0 & 0 & 12.1 & 20 & 400 \\
\underline{inhumane} & \textbf{25.7} & 1.7 & 0.6 & 3.0 & 0.7 & 90 \\
jewel   & 0.3 & 0 & \textbf{1.6} & 1.8 & 1.6 & 90 \\
\underline{karn}   & \textbf{2.3} & 0 & 1.2 & 0 & 2.1 & 170 \\
library  & \textbf{9.0} & 5.1 & 0.9 & 14.3 & 17.0 & 30 \\
ludicorp   & \textbf{10.1} & 5.4 & 8.4 & 17.8 & 13.8 & 150 \\
moonlit   & 0 & 0 & 0 & 0 & 0 & 1 \\
omniquest  & \textbf{6.9} & 4.5 & 5.6 & 3.0 & 16.8 & 50 \\
pentari   & 0 & 0 & 0 & 50.7 & 27.2 & 70 \\
\underline{snacktime}  & \textbf{19.4} & 0 & 0 & 0 & 9.7 & 50 \\
sorcerer  & \textbf{6.2} & 5.0 & 5.0 & 5.8 & 20.8 & 400 \\
\underline{spellbrkr}   & \textbf{40} & 39.9 & \textbf{40} & 21.3 & 37.8 & 600 \\
\underline{spirit}   & \textbf{1.4} & 0.6 & 1.0 & 1.3 & 0.8 & 250 \\
temple   & 0 & 0 & \textbf{7.3} & 7.6 & 7.9 & 35 \\
zenon   & 0 & 0 & 0 & 3.9 & 0 & 20 \\
zork1   & \textbf{30.4} & 24.8 & 10.3 & 34.0 & 32.6 & 350 \\
zork3   & 0.5 & 0 & \textbf{1.8} & 0.0 & 0.5 & 7 \\
ztuu  & \textbf{3.7} & 0 & 0 & 9.2 & 21.6 & 100 \\
\midrule
avg.\,norm & \textbf{9.4\%} & 5.5\% & 5.6\% & 10.8\% & 13.0\%  \\
        \end{tabular}
        
        }
    \caption{Performance of our models (\model(GPT-2) and \model(n-gram)) compared to baselines (NAIL, KG-A2C, DRRN) on Jericho. We report raw scores for individual games as well as average normalized scores (avg.\,norm). \emph{Advent} and \emph{Deephome}'s initial scores are 1 and 36, respectively. Underlined games represent those where \model outperforms handicap-assisted methods KGA2C and DRRN.}
        \label{tab:results}
\end{table}

 We provide scores of our \model-augmented DRRN agent on individual games in \tbl{tab:results}. To take into account different score scales across games, we consider both the raw score and the normalized score (raw score divided by maximum score), and only report the average normalized score across games.

Of the handicap-free models, \model (n-gram) achieves similar performance to NAIL, while \model (GPT-2) outperforms \model (n-gram) and NAIL by 4.4\% and 3.8\% on absolute normalized scores, respectively. Relatively, this represents almost a 69\% improvement over NAIL. \fig{fig:chart} presents a game-wise comparison between \model (GPT-2) and NAIL.

Surprisingly, even when compared to handicap-assisted models, \model (GPT-2) performs quite well. On 8 out of 28 games (underlined in \tbl{tab:results}), \model (GPT-2) outperforms both DRRN and KG-A2C despite the latter having access to ground-truth admissible actions. %
This improvement is especially impressive on games like \textsc{Detective}, \textsc{Inhumane} and \textsc{Snacktime}, where our normalized score is higher by more than 20\%. 
We hypothesize \model excludes some non-useful admissible actions like ``throw egg at sword'' that humans never issue, which can speed up exploration. Also, it is possible that \model sometimes discover admissible actions even the handicap cannot (due to the imperfection of state change detection).

\subsection{Analysis}
\label{sec:analysis}
\begin{figure*}[t]
    \centering
    \includegraphics[width=0.9\textwidth]{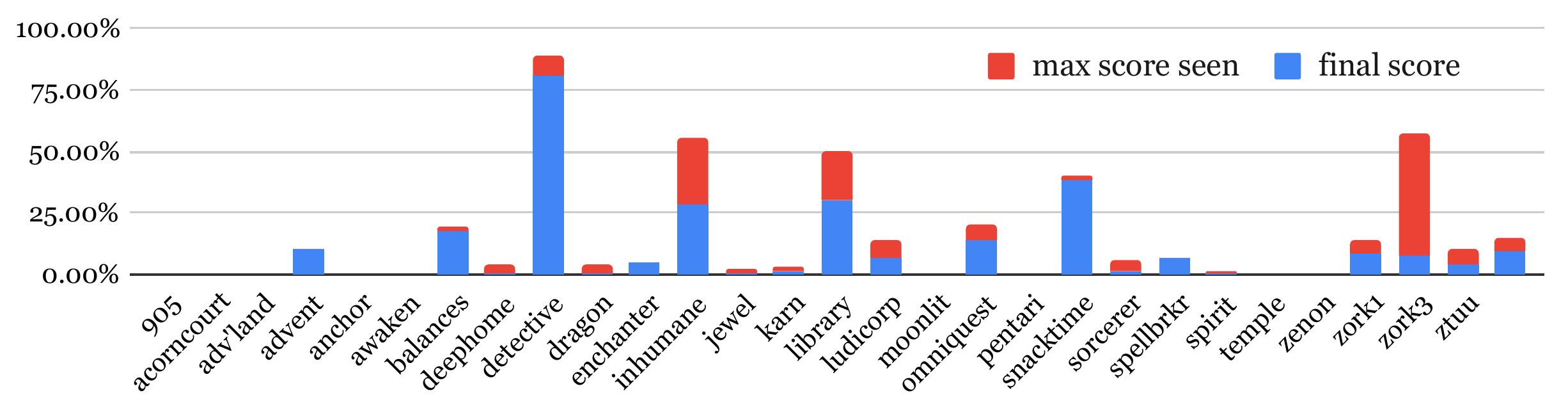}
    \caption{Final scores (blue) and maximum scores (normalized) seen during exploration (red) for \model (GPT-2). There is a lot of potential for developing better algorithms to learn from high-scoring trajectories.}
    \label{fig:seen}
\end{figure*}

\begin{table}[t]
    \centering
    \small
    \begin{tabular}{l l}
        \textbf{Variant} & \textbf{avg.\,norm} \\ \midrule
        \textit{\model (default)} & 9.4\% \\
        \midrule
        \model (20\%) & 8.1\% \\
        \model (50\%) & 8.4\% \\ \midrule
        \model (w/ Jericho) & 10.9\% \\
        \model (w/o PT) & 6.8\% \\
        \midrule
        \model ($k=10$) & 5.6\% \\
        \model ($k=20$) & 9.6\% \\
        \model ($k=40$) & 9.2\% \\ \midrule
        \model (random agent) & 1.8\% \\ 
        \end{tabular}
    \caption{Average normalized scores on Jericho for different variants of \model (GPT-2). \model (default) is the \model (GPT-2) model used for results in Table~\ref{tab:results}.
    }
        \label{tab:variants}
\end{table}

\paragraph{What Factors Contribute to Gameplay?}
We now analyze various components and design choices made in \model (GPT-2).
First, we investigate how much of the model's performance is due to pre-training on text corpora as opposed to training on our ClubFloyd data.  Then, we vary the number of actions ($k$) generated by the model. We also consider combing \model with a random agent instead of RL. This leads us to the following variants:

    1. \textbf{\model (X\%)}: These variants are trained with only X\% of the transcripts from ClubFloyd. $X=0$ is equivalent to using a pre-trained GPT-2 model off-the-shelf -- we find that this fails to produce actions that are even parseable by the game engine and therefore is not reported in the table.
    
    2. \textbf{\model (w/ Jericho)}: This variant is trained on additional ClubFloyd data that includes 8 scripts from games contained in Jericho.
    
    3. \textbf{\model (w/o PT)}: This is a randomly initialized GPT-2 model, instead of a pre-trained one, trained on ClubFloyd data. We train this model for 10 epochs until the validation loss converges, unlike previous models which we train for 3 epochs.
    
    4. \textbf{\model ($k=Y$)}: This is a model variant that produces action sets of size $Y$.
    
    5. \textbf{\model (random agent)}: This model variant replaces DRRN by a random agent that samples uniformly from \model top-30 actions at each state.
    
As shown in \tbl{tab:variants}, the significant drop in score for \model without pretraining shows that both pre-training and ClubFloyd training are important for gameplay performance. Pre-training provides general linguistic priors that regularize action generation while the ClubFloyd data conditions the model towards generating actions useful in text-based games. 

Adding heldout transcripts from Jericho evaluation games (\model w/ Jericho) provides additional benefit as expected, even surpassing handicap-assisted KG-A2C in terms of the average normalized score. Counter-intuitively, we find that the greatest performance gains aren't on games featured in the heldout transcripts. See supplementary material for more details.

For the models with different $k$ values, \model($k=10$) is much worse than other choices, but similar to \model (n-gram) in \tbl{tab:results}. Note that in \fig{fig:wt} the recall of admissible actions is similar between GPT-2 and n-gram when $k\le10$. We believe it is because top-10 GPT-2 actions are usually simple actions that occur a lot in ClubFloyd (e.g.\,`\emph{east}', `\emph{get object}'), which is also what n-gram can capture. It is really the complex actions captured when $k>10$ that makes GPT-2 much better than n-gram. On the other hand, though $k=20, 30, 40$ achieve similar overall performance, they achieve different results for different games. So potentially the \model overall performance can be further improved by choosing different $k$ for different games. Finally, \model (random agent) performs a poor score of 1.8\%, and clearly shows the importance of combining \model with an RL agent to adaptively choose actions.

\paragraph{Is \model limiting RL?}
A natural question to ask is whether reducing the action space using \model results in missing key actions that may have led to higher scores in the games. To answer this, we also plot the maximum scores seen by our \model (GPT-2) agent during RL in Figure~\ref{fig:seen}. Some games (e.g.\,\textsc{905}, \textsc{Acorncourt}) are intrinsically hard to achieve any score. However, on other games with non-zero scores, DRRN is unable to stably converge to the maximum score seen in RL exploration. If RL can fully exploit and learn from the trajectories experienced under the \model action space for each game, the average normalized score would be 14.7\%, higher than any model in \tbl{tab:results}, both with and without handicaps.

\section{Conclusion}
In this paper, we proposed the \fullmodel (\model), a language model approach to generating action candidates for reinforcement learning agents in text-based games. 
Our key insight is to use language models to capture linguistic priors and {game sense} from humans gameplay on a diverse set of games. We demonstrated that \model can generate high-quality, contextually-relevant actions even for games unseen in its training set, and when paired with a DRRN agent, outperforms previous approaches on the Jericho benchmark~\cite{hausknecht19colossal} by as much as 69\% in terms of average normalized score. Remarkably, on many of these games, our approach is competitive even with models that use ground truth admissible actions, implying that \model is able to generate high-quality actions across diverse games and contexts.

From the results in \tbl{tab:results}, it is safe to conclude that text-based games are still far from being solved. Even with access to ground truth admissible actions, sparse rewards and partial observability pose daunting challenges for current agents.
In the future, we believe that strong linguistic priors will continue to be a key ingredient for building next-level learning agents in these games. 
By releasing our dataset and code we hope to provide a solid foundation to accelerate work in this direction.

\section*{Acknowledgement}{Gracious thanks to Jacqeline Ashwell for running ClubFloyd and agreeing to our use of the collected transcripts. We thank Danqi Chen, Jimmy Yang, Jens Tuyls, and other colleagues from Princeton NLP group for proofreading and discussion. We also thank reviewers for constructive feedbacks. This research was partially funded by the Center for Statistics and Machine Learning at Princeton University through support from Microsoft.}

\bibliography{main}

\begin{thebibliography}{28}
\expandafter\ifx\csname natexlab\endcsname\relax\def\natexlab#1{#1}\fi

\bibitem[{Ammanabrolu et~al.(2019)Ammanabrolu, Broniec, Mueller, Paul, and
  Riedl}]{ammanabrolu2019toward}
Prithviraj Ammanabrolu, William Broniec, Alex Mueller, Jeremy Paul, and Mark~O
  Riedl. 2019.
\newblock Toward automated quest generation in text-adventure games.
\newblock \emph{arXiv preprint arXiv:1909.06283}.

\bibitem[{Ammanabrolu and Hausknecht(2020)}]{ammanabrolu2020graph}
Prithviraj Ammanabrolu and Matthew Hausknecht. 2020.
\newblock Graph constrained reinforcement learning for natural language action
  spaces.
\newblock \emph{arXiv preprint arXiv:2001.08837}.

\bibitem[{Ammanabrolu and Riedl(2019)}]{ammanabrolu2019playing}
Prithviraj Ammanabrolu and Mark Riedl. 2019.
\newblock Playing text-adventure games with graph-based deep reinforcement
  learning.
\newblock In \emph{Proceedings of the 2019 Conference of the North American
  Chapter of the Association for Computational Linguistics: Human Language
  Technologies, Volume 1 (Long and Short Papers)}, pages 3557--3565.

\bibitem[{Chen et~al.(2017)Chen, Liu, Yin, and Tang}]{chen2017survey}
Hongshen Chen, Xiaorui Liu, Dawei Yin, and Jiliang Tang. 2017.
\newblock A survey on dialogue systems: Recent advances and new frontiers.
\newblock \emph{Acm Sigkdd Explorations Newsletter}, 19(2):25--35.

\bibitem[{Cho et~al.(2014)Cho, van Merri{\"e}nboer, Bahdanau, and
  Bengio}]{cho-etal-2014-properties}
Kyunghyun Cho, Bart van Merri{\"e}nboer, Dzmitry Bahdanau, and Yoshua Bengio.
  2014.
\newblock On the properties of neural machine translation: Encoder{--}decoder
  approaches.
\newblock In \emph{Proceedings of {SSST}-8, Eighth Workshop on Syntax,
  Semantics and Structure in Statistical Translation}, pages 103--111, Doha,
  Qatar. Association for Computational Linguistics.

\bibitem[{C{\^o}t{\'e} et~al.(2018)C{\^o}t{\'e}, K{\'a}d{\'a}r, Yuan, Kybartas,
  Barnes, Fine, Moore, Hausknecht, El~Asri, Adada et~al.}]{cote2018textworld}
Marc-Alexandre C{\^o}t{\'e}, {\'A}kos K{\'a}d{\'a}r, Xingdi Yuan, Ben Kybartas,
  Tavian Barnes, Emery Fine, James Moore, Matthew Hausknecht, Layla El~Asri,
  Mahmoud Adada, et~al. 2018.
\newblock Textworld: A learning environment for text-based games.
\newblock In \emph{Workshop on Computer Games}, pages 41--75. Springer.

\bibitem[{Devlin et~al.(2018)Devlin, Chang, Lee, and
  Toutanova}]{devlin2018bert}
Jacob Devlin, Ming-Wei Chang, Kenton Lee, and Kristina Toutanova. 2018.
\newblock Bert: Pre-training of deep bidirectional transformers for language
  understanding.
\newblock \emph{arXiv preprint arXiv:1810.04805}.

\bibitem[{Fulda et~al.(2017)Fulda, Ricks, Murdoch, and Wingate}]{fulda17}
Nancy Fulda, Daniel Ricks, Ben Murdoch, and David Wingate. 2017.
\newblock What can you do with a rock? affordance extraction via word
  embeddings.
\newblock \emph{CoRR}, abs/1703.03429.

\bibitem[{Hausknecht et~al.(2019{\natexlab{a}})Hausknecht, Ammanabrolu,
  C{\^{o}}t{\'{e}}, and Yuan}]{hausknecht19colossal}
Matthew Hausknecht, Prithviraj Ammanabrolu, Marc-Alexandre C{\^{o}}t{\'{e}},
  and Xingdi Yuan. 2019{\natexlab{a}}.
\newblock Interactive fiction games: A colossal adventure.
\newblock \emph{CoRR}, abs/1909.05398.

\bibitem[{Hausknecht et~al.(2019{\natexlab{b}})Hausknecht, Loynd, Yang,
  Swaminathan, and Williams}]{hausknecht2019nail}
Matthew Hausknecht, Ricky Loynd, Greg Yang, Adith Swaminathan, and Jason~D
  Williams. 2019{\natexlab{b}}.
\newblock Nail: A general interactive fiction agent.
\newblock \emph{arXiv preprint arXiv:1902.04259}.

\bibitem[{He et~al.(2015)He, Chen, He, Gao, Li, Deng, and
  Ostendorf}]{he2015deep}
Ji~He, Jianshu Chen, Xiaodong He, Jianfeng Gao, Lihong Li, Li~Deng, and Mari
  Ostendorf. 2015.
\newblock Deep reinforcement learning with a natural language action space.
\newblock \emph{arXiv preprint arXiv:1511.04636}.

\bibitem[{Jain et~al.(2019)Jain, Fedus, Larochelle, Precup, and
  Bellemare}]{jain2019}
Vishal Jain, William Fedus, Hugo Larochelle, Doina Precup, and Marc~G.
  Bellemare. 2019.
\newblock Algorithmic improvements for deep reinforcement learning applied to
  interactive fiction.
\newblock \emph{CoRR}, abs/1903.03094.

\bibitem[{Joulin et~al.(2017)Joulin, Grave, Bojanowski, and
  Mikolov}]{joulin2017bag}
Armand Joulin, Edouard Grave, Piotr Bojanowski, and Tomas Mikolov. 2017.
\newblock Bag of tricks for efficient text classification.
\newblock In \emph{Proceedings of the 15th Conference of the European Chapter
  of the Association for Computational Linguistics: Volume 2, Short Papers},
  pages 427--431. Association for Computational Linguistics.

\bibitem[{Jurafsky and Martin(2009)}]{JurafskyMartin09}
Daniel Jurafsky and James~H. Martin. 2009.
\newblock \emph{Speech and Language Processing (2nd Edition)}.
\newblock Prentice-Hall, Inc., USA.

\bibitem[{Kostka et~al.(2017)Kostka, Kwiecien, Kowalski, and
  Rychlikowski}]{kostka17}
Bartosz Kostka, Jaroslaw Kwiecien, Jakub Kowalski, and Pawel Rychlikowski.
  2017.
\newblock Text-based adventures of the golovin {AI} agent.
\newblock \emph{CoRR}, abs/1705.05637.

\bibitem[{Lazaridou et~al.(2020)Lazaridou, Potapenko, and
  Tieleman}]{lazaridou2020multi}
Angeliki Lazaridou, Anna Potapenko, and Olivier Tieleman. 2020.
\newblock Multi-agent communication meets natural language: Synergies between
  functional and structural language learning.
\newblock \emph{arXiv preprint arXiv:2005.07064}.

\bibitem[{Mikolov et~al.(2013)Mikolov, Sutskever, Chen, Corrado, and
  Dean}]{mikolov13}
Tomas Mikolov, Ilya Sutskever, Kai Chen, Greg~S Corrado, and Jeff Dean. 2013.
\newblock Distributed representations of words and phrases and their
  compositionality.
\newblock In C.~J.~C. Burges, L.~Bottou, M.~Welling, Z.~Ghahramani, and K.~Q.
  Weinberger, editors, \emph{Advances in Neural Information Processing Systems
  26}, pages 3111--3119. Curran Associates, Inc.

\bibitem[{Narasimhan et~al.(2015)Narasimhan, Kulkarni, and
  Barzilay}]{narasimhan15}
Karthik Narasimhan, Tejas~D. Kulkarni, and Regina Barzilay. 2015.
\newblock Language understanding for text-based games using deep reinforcement
  learning.
\newblock In \emph{EMNLP}, pages 1--11.

\bibitem[{Radford et~al.(2019)Radford, Wu, Child, Luan, Amodei, and
  Sutskever}]{radford2019language}
Alec Radford, Jeffrey Wu, Rewon Child, David Luan, Dario Amodei, and Ilya
  Sutskever. 2019.
\newblock Language models are unsupervised multitask learners.
\newblock \emph{OpenAI Blog}, 1(8):9.

\bibitem[{Song et~al.(2016)Song, Yan, Li, Zhao, and Zhang}]{song2016two}
Yiping Song, Rui Yan, Xiang Li, Dongyan Zhao, and Ming Zhang. 2016.
\newblock Two are better than one: An ensemble of retrieval-and
  generation-based dialog systems.
\newblock \emph{arXiv preprint arXiv:1610.07149}.

\bibitem[{Sutskever et~al.(2014)Sutskever, Vinyals, and
  Le}]{sutskever2014sequence}
Ilya Sutskever, Oriol Vinyals, and Quoc~V Le. 2014.
\newblock Sequence to sequence learning with neural networks.
\newblock In \emph{Advances in neural information processing systems}, pages
  3104--3112.

\bibitem[{Tao et~al.(2018)Tao, C{\^o}t{\'e}, Yuan, and Asri}]{tao2018towards}
Ruo~Yu Tao, Marc-Alexandre C{\^o}t{\'e}, Xingdi Yuan, and Layla~El Asri. 2018.
\newblock Towards solving text-based games by producing adaptive action spaces.
\newblock \emph{arXiv preprint arXiv:1812.00855}.

\bibitem[{Urbanek et~al.(2019)Urbanek, Fan, Karamcheti, Jain, Humeau, Dinan,
  Rocktäschel, Kiela, Szlam, and Weston}]{urbanek2019light}
Jack Urbanek, Angela Fan, Siddharth Karamcheti, Saachi Jain, Samuel Humeau,
  Emily Dinan, Tim Rocktäschel, Douwe Kiela, Arthur Szlam, and Jason Weston.
  2019.
\newblock Learning to speak and act in a fantasy text adventure game.
\newblock \emph{CoRR}, abs/1903.03094.

\bibitem[{Walton(2019)}]{walton}
Nick Walton. 2019.
\newblock Ai dungeon 2: Creating infinitely generated text adventures with deep
  learning language models.

\bibitem[{Williams et~al.(2017)Williams, Asadi, and
  Zweig}]{williams-etal-2017-hybrid}
Jason~D. Williams, Kavosh Asadi, and Geoffrey Zweig. 2017.
\newblock Hybrid code networks: practical and efficient end-to-end dialog
  control with supervised and reinforcement learning.
\newblock In \emph{Proceedings of the 55th Annual Meeting of the Association
  for Computational Linguistics (Volume 1: Long Papers)}, pages 665--677,
  Vancouver, Canada. Association for Computational Linguistics.

\bibitem[{Wolf et~al.(2019)Wolf, Debut, Sanh, Chaumond, Delangue, Moi, Cistac,
  Rault, Louf, Funtowicz, and Brew}]{Wolf2019HuggingFacesTS}
Thomas Wolf, Lysandre Debut, Victor Sanh, Julien Chaumond, Clement Delangue,
  Anthony Moi, Pierric Cistac, Tim Rault, R'emi Louf, Morgan Funtowicz, and
  Jamie Brew. 2019.
\newblock Huggingface's transformers: State-of-the-art natural language
  processing.
\newblock \emph{ArXiv}, abs/1910.03771.

\bibitem[{Zahavy et~al.(2018)Zahavy, Haroush, Merlis, Mankowitz, and
  Mannor}]{zahavy18}
Tom Zahavy, Matan Haroush, Nadav Merlis, Daniel~J Mankowitz, and Shie Mannor.
  2018.
\newblock Learn what not to learn: Action elimination with deep reinforcement
  learning.
\newblock In S.~Bengio, H.~Wallach, H.~Larochelle, K.~Grauman, N.~Cesa-Bianchi,
  and R.~Garnett, editors, \emph{Advances in Neural Information Processing
  Systems 31}, pages 3562--3573. Curran Associates, Inc.

\bibitem[{Zhao and Eskenazi(2016)}]{zhao-eskenazi-2016-towards}
Tiancheng Zhao and Maxine Eskenazi. 2016.
\newblock Towards end-to-end learning for dialog state tracking and management
  using deep reinforcement learning.
\newblock In \emph{Proceedings of the 17th Annual Meeting of the Special
  Interest Group on Discourse and Dialogue}, pages 1--10, Los Angeles.
  Association for Computational Linguistics.

\end{thebibliography}
\bibliographystyle{acl_natbib}

\appendix
\section{ClubFloyd Dataset}
\label{appendix:clubfloyd}
\begin{figure}[t]
\centering
\includegraphics[width=.49\textwidth]{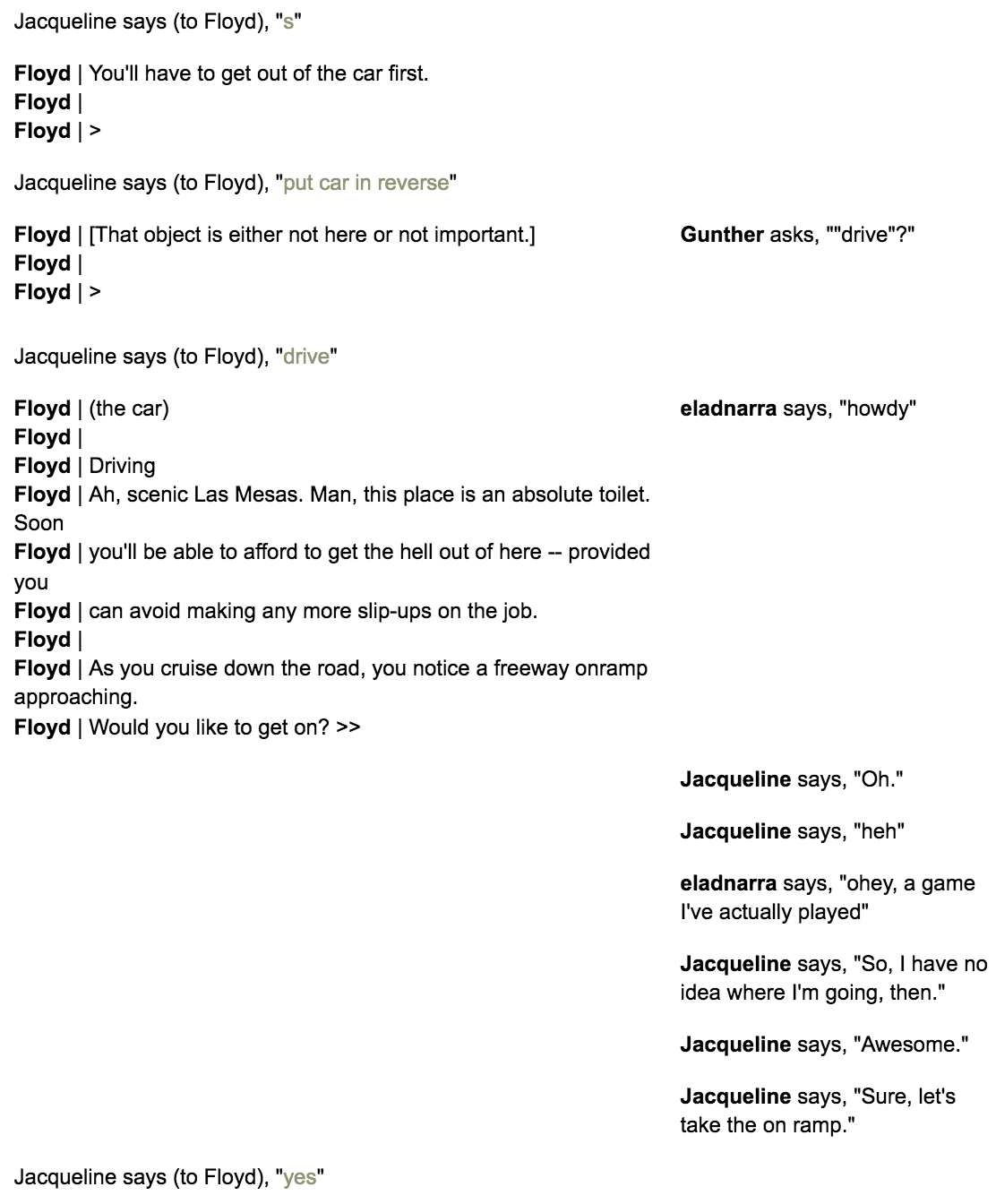}
\caption{Selection from a raw ClubFloyd Transcript of the game 9:05}
\label{fig:clubfloyd_cleaning}
\end{figure}

The ClubFloyd transcripts we collected are gameplay logs generated by a group of people that regularly meet to play interactive fiction games. The participants are experienced at playing text-based games, however they may not be familiar with the game that's being played, and do make several mistakes. We include a snippet of a transcript in Figure \ref{fig:clubfloyd_cleaning}. We crawled the ClubFloyd website to acquire 426 transcripts, spanning over 500 games.

To process a transcript, we clean the data and extract observations and actions. The data contains several sources of noise, which we remove: the first is non-game information such as chat logs between the humans playing the games; second are meta-actions that humans use to save and load games and navigate menus; and finally, we remove typos, expand common abbreviations (``n" to ``north", ``x" to ``examine", etc.), and filter out any actions that weren't recognized by the game parsers. 

\begin{figure}[ht]
\begin{mdframed}
  \parbox{\textwidth}{

\tiny [OBS] [That object is either not here or not important.] [ACTION] south [OBS] You'll have to get out of the car first. [ACTION] \textbf{put car in reverse}\\

[OBS] You'll have to get out of the car first. [ACTION] put car in reverse [OBS] [That object is either not here or not important.] [ACTION] \textbf{drive}\\

[OBS] [That object is either not here or not important.] [ACTION] drive [OBS] (the car) Driving Ah, scenic Las Mesas. Man, this place is an absolute toilet. Soon you'll be able to afford to get the hell out of here -- provided you can avoid making any more slip-ups on the job. As you cruise down the road, you notice a freeway onramp approaching. Would you like to get on? $>>$ [ACTION] \textbf{yes}

}
\end{mdframed}
\caption{Cleaned section of Figure \ref{fig:clubfloyd_cleaning}}
    \label{cf-ex}
 \end{figure}

Once we have our cleaned observations and actions, we group observations and actions into the form $(o_{j-1}, a_{j-1}, o_j), a_j$. For the very first observation and action, we pad the beginning of the example with the observation "You are at the start of your journey" and the action "begin journey". 

After this entire pre-processing, the dataset contains 223,527 examples.

\section{\model Training}
In this section, we will provide training details of \model(GPT-2), \model(n-gram), and their variants. 
\subsection{\model(GPT-2)} We first discuss the \model(GPT-2) models, and begin with the portion of the ClubFloyd data that they are trained on. We begin with a 12-layer, 768-hidden, 12-head, 117M parameter pretrained OpenAI GPT-2 model.

We note that the number of samples we train on, even in the \model(GPT-2) model + Jericho games variant, is less than the total samples in the dataset. This is because we do not train on incomplete batches of data, and we omit samples that exceed 256 tokens.

\paragraph{\model(GPT-2)} To train \model(GPT-2), we take transcripts from ClubFloyd (excluding Jericho games) and order the samples based on the transcript number they came from. This yields a dataset of 193,588 samples. We select the first 90\% of the samples as train data, and the last 10\% of the samples as validation data. 
\paragraph{\model(GPT-2) 50\%, 20\%, (+) Jericho} To train the 50\% and 20\% variants, we select without replacement 212 transcripts (94,609 samples), and 85 transcripts (38,334 samples) respectively from the ClubFloyd transcripts (excluding Jericho games). We order the samples based on the transcript they come from, choose the first 90\% of the data as our training data and last 10\% as validation data.

For the \model(GPT-2) variant including Jericho games, we include every ClubFloyd transcript, we randomly order the transcripts, order the samples based on the order of the transcripts, and then we select the first 90\% of the data as our training data, and the last 10\% of the data as validation data. This split contains 206,286 samples.

\paragraph{\model(GPT-2) Random Initialization} For the \model(GPT-2) variant with random initialization, we begin with a GPT-2 model that has not been pretrained. We only use the transcripts in ClubFloyd that do not correspond to any Jericho game we test on. We randomly order the transcripts, and order the samples based on the order of the transcripts. We select the first 90\% of the data as our training data, and the last 10\% of the data as validation data.

\paragraph{Parameter Optimization}
In order to train GPT-2, we minimize the cross-entropy between GPT-2's distribution over actions and the action taken in the ClubFloyd example. We use Adam to optimize the weights of our model with learning rate = 2e-5 and Adam epsilon = 1e-8. For the learning rate we use a linear schedule with warmup. Finally, we clip gradients allowing a max gradient norm of 1. 

We include the loss on the train and validation set, as well as the accuracy (defined as the percentage of examples on which the action assigned the highest probability by GPT-2 was the ClubFloyd action) in Table \ref{tab:training_stats}.
\begin{table}[t]
    \centering
    \small
\resizebox{\columnwidth}{!}{
    \begin{tabular}{cc|ccccccc}
           Model & Metric &  1  &  2  & 3  & 4  &  5  &  9 &  10\\
        \midrule
        Main & Train Loss & 0.32 & 0.27 & 0.25 & 0.23 & 0.22 & n/a & n/a \\
         & Train Acc & 0.11 & 0.14 & 0.16 & 0.18 & 0.19 & n/a & n/a \\
         & Val Loss & 2.14 & 2.04 & 1.98 & 1.96 & 1.96 & n/a & n/a \\
         & Val Acc & 0.13 & 0.15 & 0.16 & 0.17 & 0.18 & n/a & n/a \\
        \midrule
        50\% & Train Loss & 0.66 & 0.55 & 0.49 & 0.46 & 0.43 & n/a & n/a \\
        & Train Acc & 0.11 & 0.14 & 0.17 & 0.19 & 0.21 & n/a & n/a \\
        & Val Loss & 2.19 & 2.09 & 2.06 & 2.04 & 2.05 & n/a & n/a \\
        & Val Acc & 0.14 & 0.15 & 0.15 & 0.16 & 0.16 & n/a & n/a \\
        \midrule
        20\% & Train Loss & 0.37 & 0.29 & 0.26 & 0.25 & 0.24 & n/a & n/a \\
        & Train Acc & 0.08 & 0.11 & 0.13 & 0.15 & 0.16 & n/a & n/a \\
        & Val Loss & 2.32 & 2.17 & 2.12 & 2.09 & 2.08 & n/a & n/a \\
         & Val Acc & 0.10 & 0.12 & 0.13 & 0.14 & 0.15 & n/a & n/a \\
        \midrule
        Jericho  & Train Loss & 0.62 & 0.53 & 0.48 & 0.45 & 0.43 & n/a & n/a \\
        & Train Acc & 0.12 & 0.16 & 0.19 & 0.21 & 0.23 & n/a & n/a \\
        & Val Loss & 2.10 & 2.00 & 1.97 & 1.96 & 1.98 & n/a & n/a \\
         & Val Acc & 0.16 & 0.17 & 0.17 & 0.18 & 0.18 & n/a & n/a \\
        \midrule
        Random Init& Train Loss & 0.36 & 0.33 & 0.31 & 0.29 & 0.27 & 0.23 & 0.23 \\
         & Train Acc & 0.05 & 0.07 & 0.08 & 0.10 & 0.11 & 0.15 & 0.15 \\
         & Val Loss & 4.96 & 4.60 & 4.35 & 4.16 & 4.01 & 3.73 & 3.73 \\
        & Val Acc & 0.06 & 0.08 & 0.09 & 0.10 & 0.10 & 0.12 & 0.12 \\
        \end{tabular}
        }
    \caption{Training Metrics for CALM Variants}
        \label{tab:training_stats}
\end{table}

\subsection{\model(n-gram)} In order to train the \model n-gram model, we consider the set of transcripts in ClubFloyd (excluding Jericho games). Next, we take the set of actions that appear in these transcripts, and train an n-gram model with Laplace $\alpha$ smoothing to model these sequences \cite{JurafskyMartin09}. We order actions by the transcript they appear in and take the first 70\% of the actions as train data and leave the remaining 30\% as validation data. For each $n$, we choose alpha that minimizes perplexity per word on the validation data. We also tried a linear interpolation of these estimates \cite{JurafskyMartin09} although we did not observe an improvement over our bigram model. In this model, we estimate $p(a^i|a^{i-3},a^{i-2}, a^{i-1}) = w_1p^*(a^i|a^{i-3},a^{i-2}, a^{i-1}) + w_2p^*(a^i|a^{i-2}, a^{i-1}) + w_3p^*(a^i| a^{i-1}) + w_4 p^*(a^i)$
where $\sum_i w_i = 1$, and $p^*$ indicates our m-gram estimate for $p(a^i|a^{i-m+1},..., a^{i-1})$.

\section{Walkthrough Evaluation}
In \fig{fig:wt_zork1}, we provide a piece of walkthrough trajectory of Zork1, with GPT-2 and n-gram generated actions at each state. Note that n-gram actions are mostly limited to be no more than two tokens, while GPT-2 can generate more complex actions like ``put sword in case''. 

\begin{figure*}[t]
    \centering
    \includegraphics[width=\textwidth]{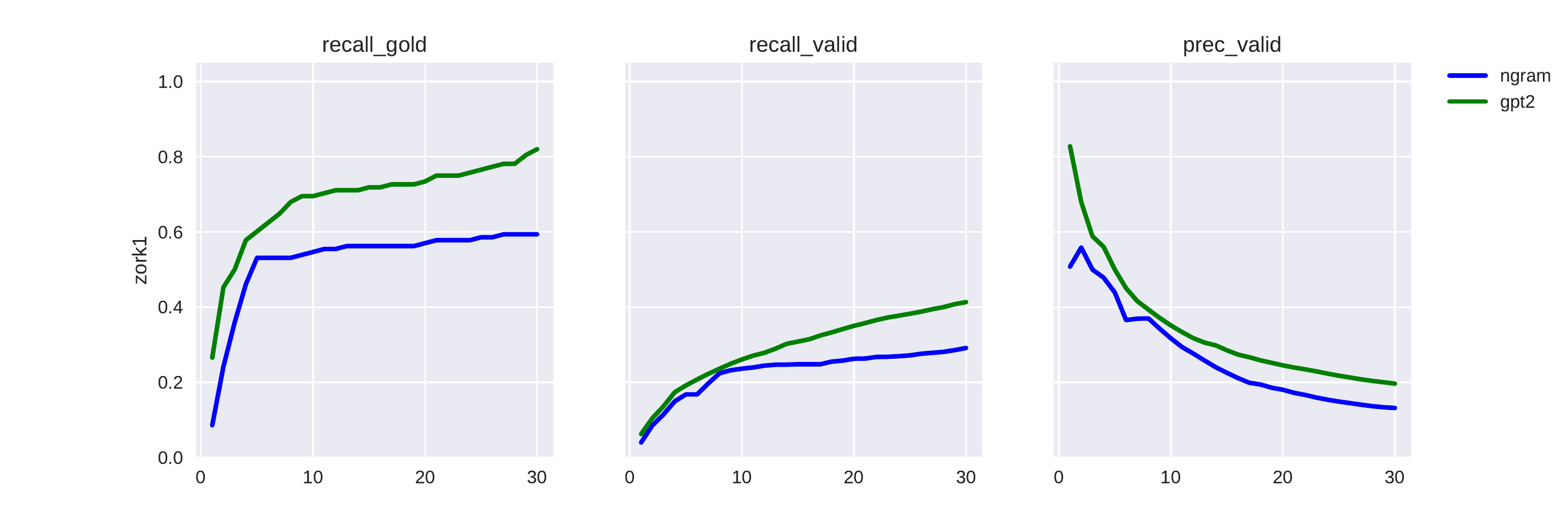}
        \includegraphics[width=\textwidth]{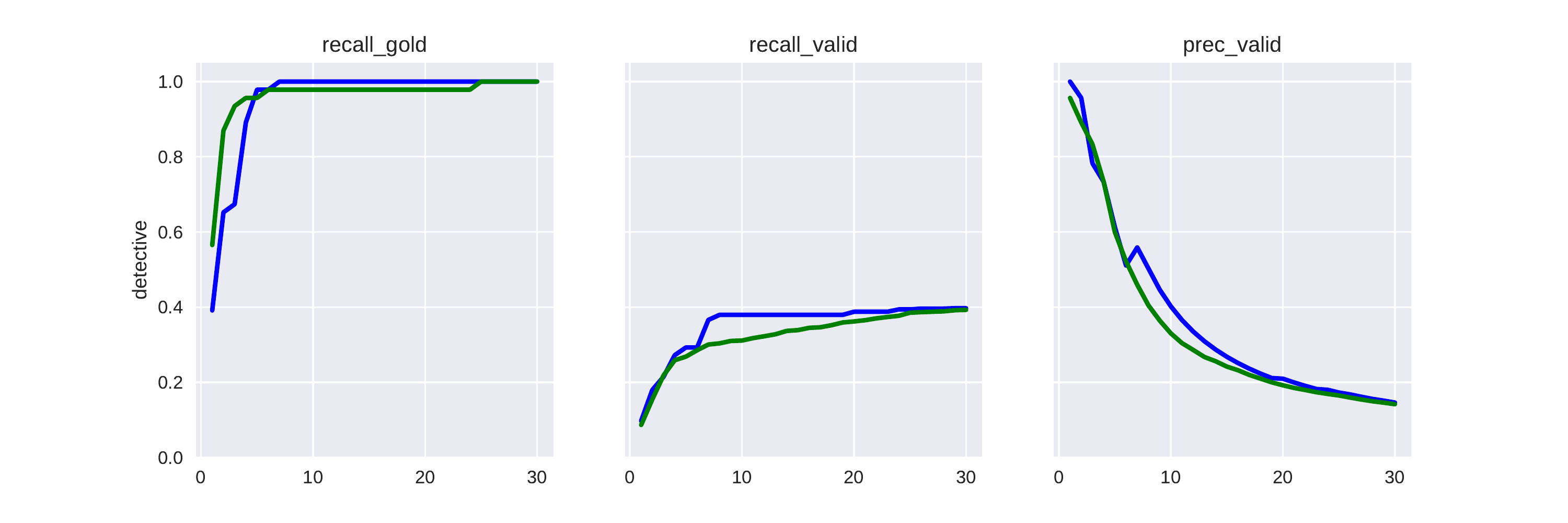}
    \caption{Walkthrough evaluation for Zork1 and Detective.}
    \label{fig:wt-eval}
\end{figure*}

In \fig{fig:wt-eval}, we provide game-specific metric curves for Zork1 and Detective. On harder games like Zork1, there is significant gap between GPT-2 and n-gram, while easy games like Detective the gap is very small.

\section{Gameplay Evaluation}
\begin{figure*}
    \centering
    \includegraphics[width=\textwidth]{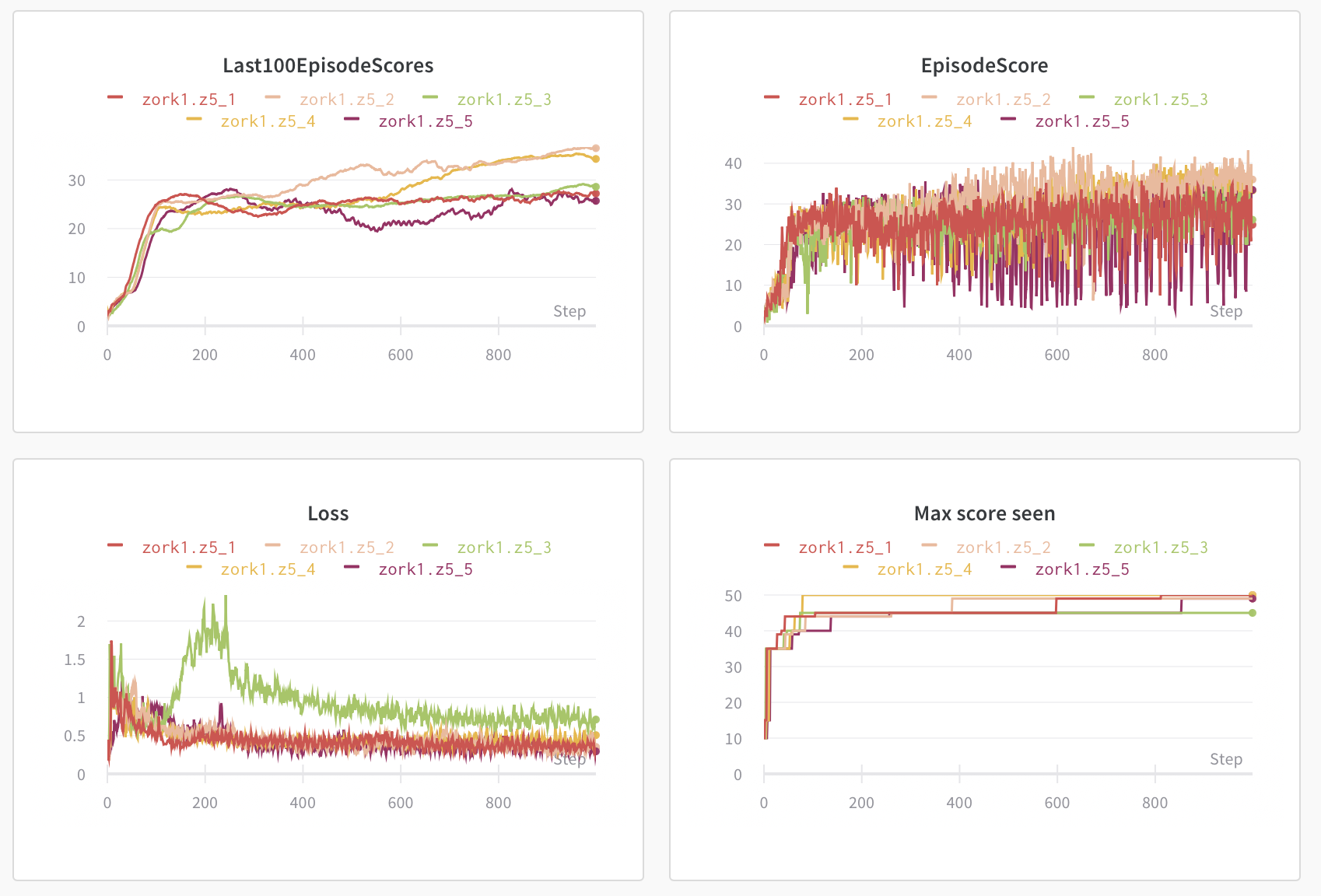}
    \caption{\model (GPT-2) learning Zork1. Results show the five independent training runs.}
    \label{fig:zork1-rl}
\end{figure*}

\begin{figure*}
    \centering
    \includegraphics[width=\textwidth]{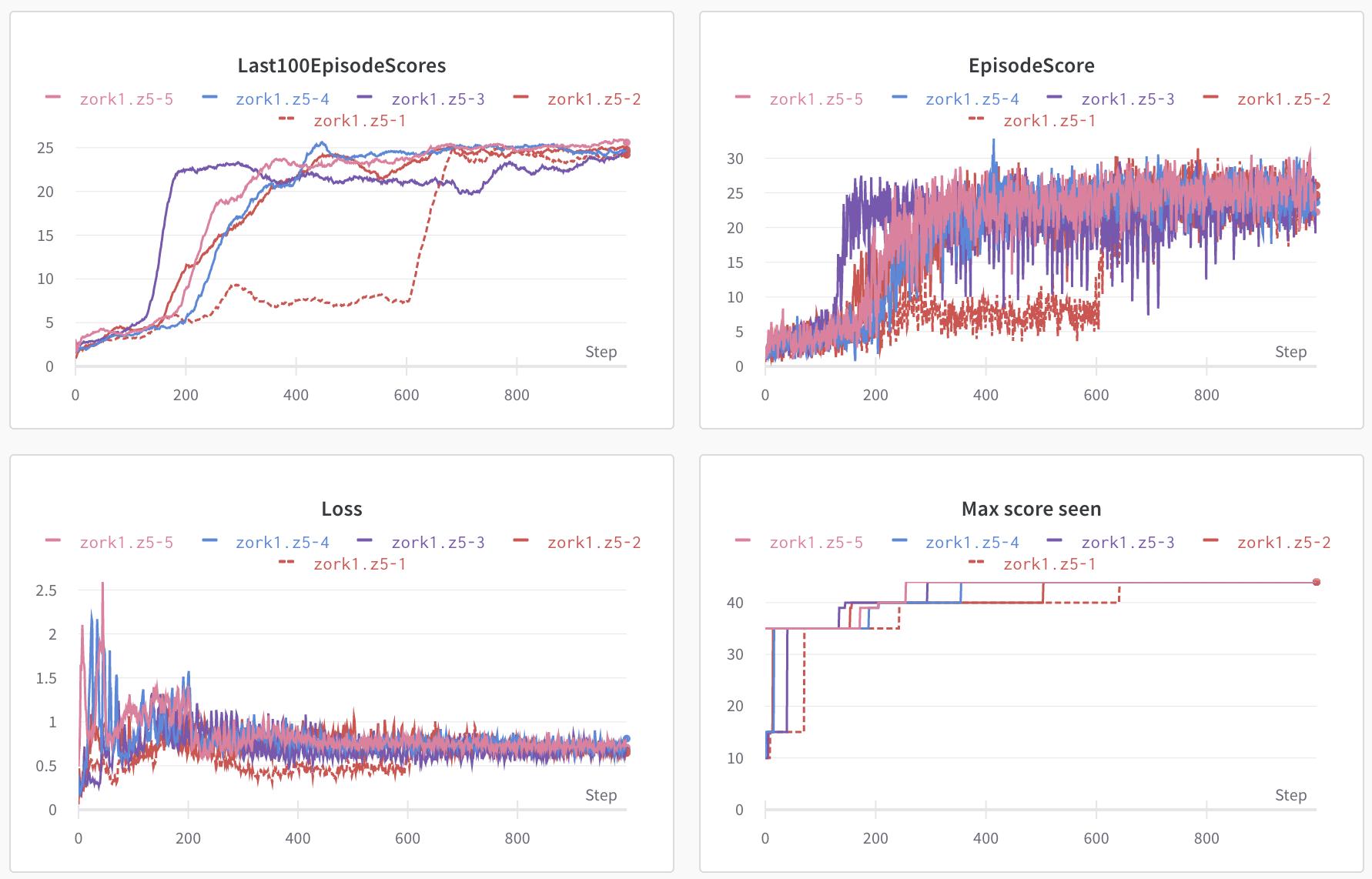}
    \caption{\model (n-gram) learning Zork1. Results show the five independent training runs.}
        \label{fig:zork1-rl-ngram}
\end{figure*}

\begin{figure*}
    \centering
    \includegraphics[width=\textwidth]{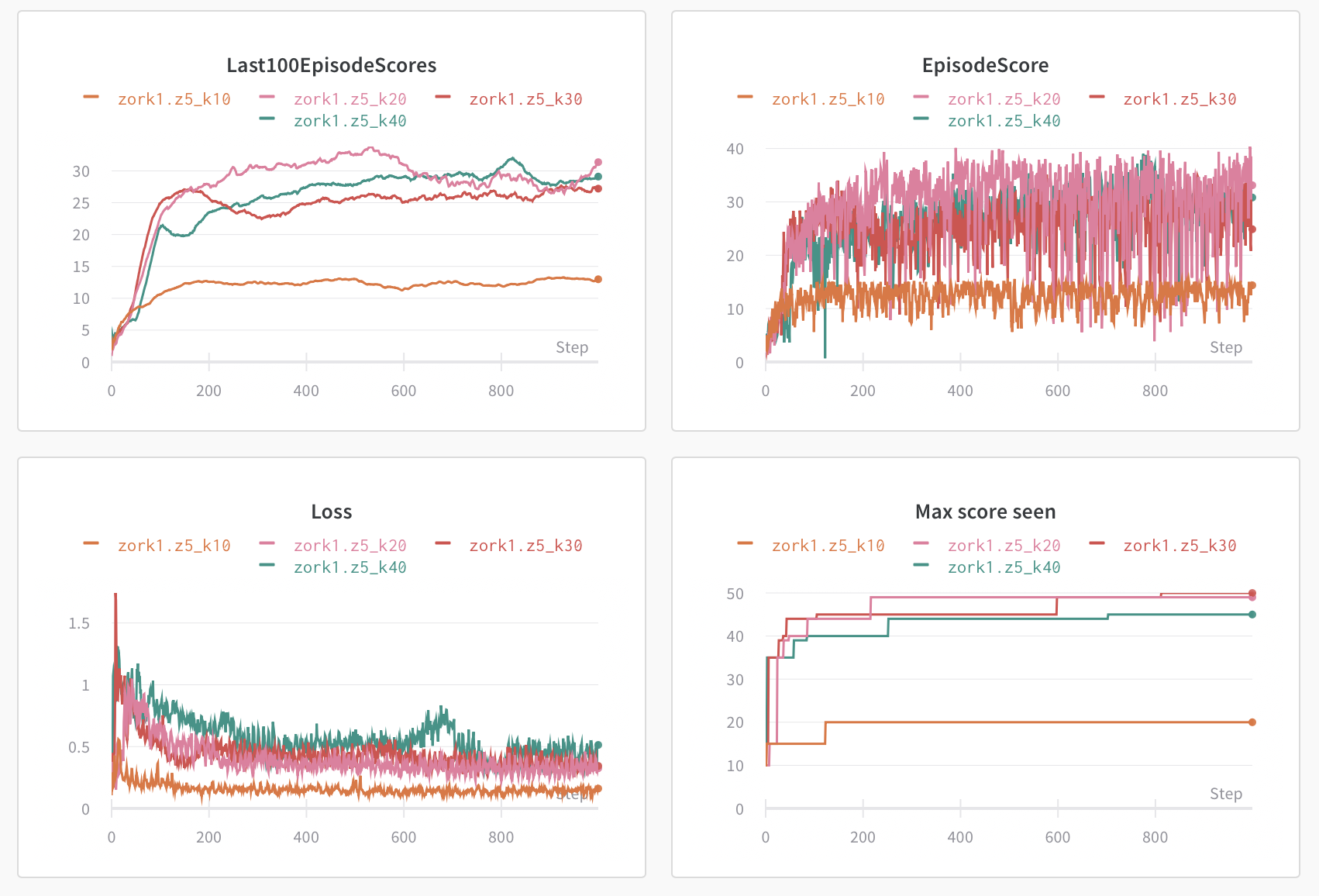}
    \caption{\model (GPT-2) on Zork1 when decoding variable numbers of top-$k$ actions ($k=10,20,30,40$).}
    \label{fig:zork1-rl-k}
\end{figure*}

\begin{figure*}
    \centering
    \includegraphics[width=\textwidth]{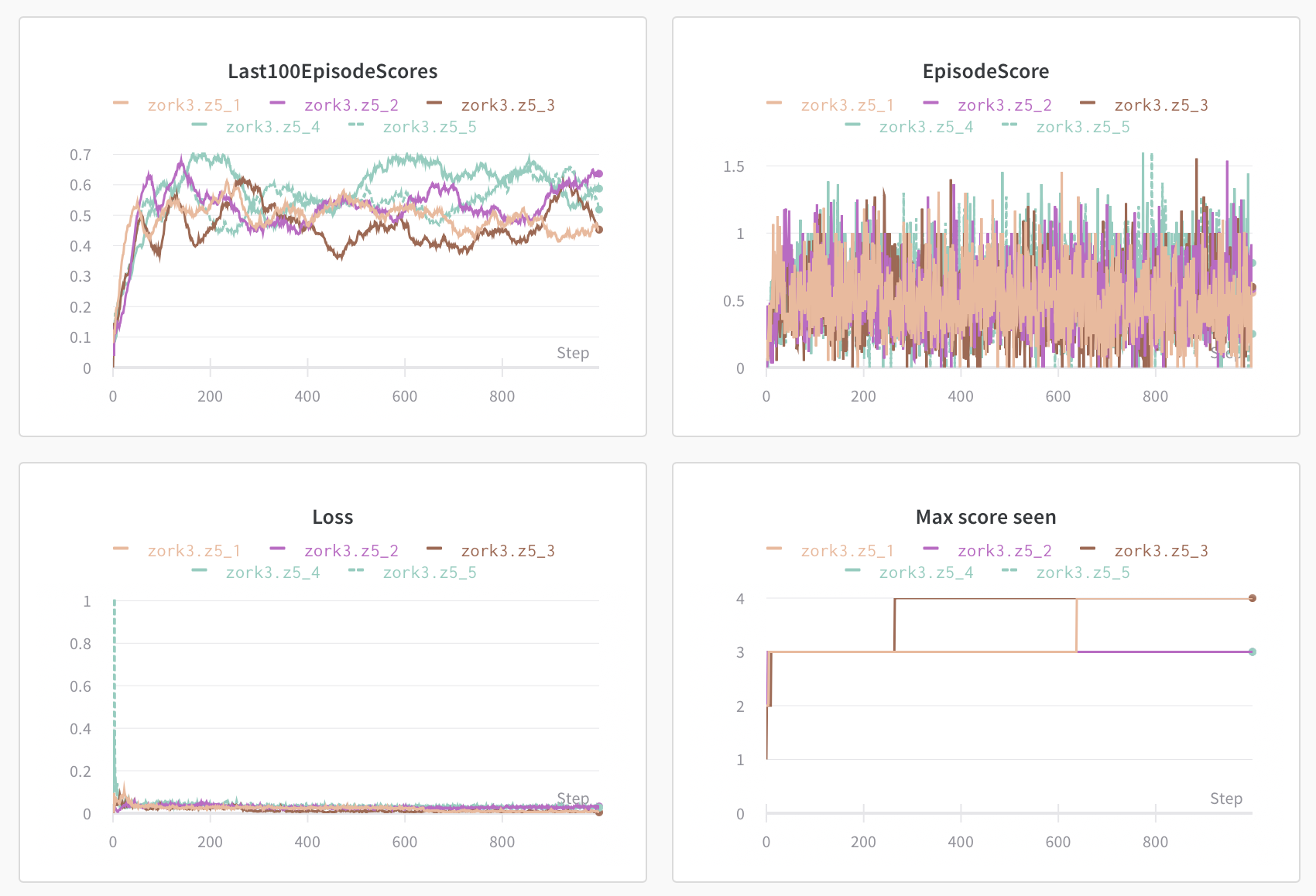}
    \caption{\model (GPT-2) learning curves on Zork3.}
    \label{fig:zork3-rl}
\end{figure*}

On Zork1, we provide learning curves for \model (GPT-2) (\fig{fig:zork1-rl}) and \model (n-gram) (\fig{fig:zork1-rl-ngram}). We also provide trail curves for \model (GPT-2) on Zork3 (\fig{fig:zork3-rl}), a game we are behind NAIL, and trails using different top-$k \in \{10, 20, 30, 40\}$ actions by \model (GPT-2) on Zork1 (\fig{fig:zork1-rl-k}). 

We provide per-game results for model variants in \tbl{tab:variant-pergame}. It is interesting that \model (w/ Jericho) is significantly better than \model (GPT-2) on the games of Temple and Deephome (non-trivial scores achieved), which are not the games with ClubFloyd scripts added. On the other hand, games like 905 and moonlit have scripts added, but do not get improved. 

In the end, we append one example trajectory piece of DRRN + \model (GPT-2) on Zork1 (\fig{fig:end}), where \model generated action candidates and their Q-values are shown along with observations, actions and scores.

\begin{table*}[t]
    \centering
\resizebox{2\columnwidth}{!}{
    \begin{tabular}{r|l|l|llllllll|l}
Game  & CALM (GPT-2) &  CALM (ngram) & CALM (w/o PT) & CALM (20\%) & CALM (50\%) & CALM (w/ Jericho) & CALM (k=10) & CALM (k=20) & CALM (k=40) & CALM (random agent) &  Max Score \\ \midrule
\textbf{905} & 0.00 ($\pm$ 0.00) &  0.00 ($\pm$ 0.00) &0.00 & 0.00 & 0.00 & 0.00 ($\pm$ 0.00) & 0.00 & 0.00 & 0.00 &0.00 & 1 \\
acorncourt   & 0.00 ($\pm$ 0.00) & 0.00 ($\pm$ 0.00) &0.00 & 0.00 & 0.00 & 0.00 ($\pm$ 0.00) & 0.00 & 0.00 & 0.00 & 0.00 & 30 \\
adv'land   & 0.00 ($\pm$ 0.00) &0.00 ($\pm$ 0.00)& 0.00 & 0.00 & 0.00 & 0.00 ($\pm$ 0.00) & 0.00 & 0.00 & 0.00 & 0.00 & 100 \\
advent & 36.00 ($\pm$ 0.00) &  36.00 ($\pm$ 0.00)&36.00 & 36.00 & 36.00 & 36.00 ($\pm$ 0.00) & 36.00 & 36.00 & 36.00 & 36.00 & 350 \\
anchor   & 0.00 ($\pm$ 0.00) &0.00 ($\pm$ 0.00)& 0.00 & 0.00 & 0.00 & 0.00 ($\pm$ 0.00) & 0.00 & 0.00 & 0.00 &  0.00 & 100 \\
awaken  & 0.00 ($\pm$ 0.00) & 0.00 ($\pm$ 0.00)& 0.00 & 0.00 & 0.00 & 0.00 ($\pm$ 0.00) & 0.00 & 0.00 & 0.00 & 0.00 & 50 \\
balances  & 9.15 ($\pm$ 0.08) & 8.86 ($\pm$ 0.04)& 6.00 & 7.89 & 9.43 & 4.05 ($\pm$ 0.15) & 0.00 & 9.17 & 8.07 & 1.70 & 51 \\
deephome   & 1.00 ($\pm$ 0.00) &1.00 ($\pm$ 0.00)& 1.00 & 1.00 & 1.00 & 6.95 ($\pm$ 5.43) & 1.00 & 1.00 & 1.00 & 1.05 & 300 \\
detective  & 289.71 ($\pm$ 0.20) &284.33 ($\pm$ 11.04) & 288.21 & 289.30 & 289.58 & 289.87 ($\pm$ 0.11) & 289.75 & 289.51 & 290.04 & 40.00 & 360 \\
dragon   & 0.13 ($\pm$ 0.05) & 0.05 ($\pm$ 0.03)&0.00 & 0.27 & 0.25 & 0.19 ($\pm$ 0.03) & 0.32 & 0.12 & 0.18 &  -0.19 & 25 \\
\textbf{enchanter}  & 19.09 ($\pm$ 0.59) & 0.00 ($\pm$ 0.00)& 0.00 & 0.00 & 0.00 & 19.92 ($\pm$ 0.06) & 0.00 & 15.33 & 20.00 &0.00 & 400 \\
inhumane   & 25.73 ($\pm$ 2.93) & 1.72 ($\pm$ 0.93) & 0.00 & 20.15 & 22.38 & 28.16 ($\pm$ 3.32) & 8.38 & 30.03 & 21.73& 0.00 & 90 \\
jewel   & 0.27 ($\pm$ 0.01) & 0.00 ($\pm$ 0.00) & 0.00 & 0.00 & 0.00 & 0.38 ($\pm$ 0.05) & 0.00 & 0.20 & 0.46 & 0.00 & 90 \\
karn   & 2.30 ($\pm$ 0.05) &0.00 ($\pm$ 0.00) &  0.00 & 3.19 & 1.73 & 2.19 ($\pm$ 0.08) & 0.14 & 2.63 & 1.71 &0.00 & 170 \\
library  & 9.02 ($\pm$ 5.07) &5.07 ($\pm$ 0.28)& 13.77 & 12.31 & 11.84 & 12.47 ($\pm$ 0.35) & 3.22 & 10.40 & 10.46 & 1.74 & 30 \\
ludicorp   & 10.09 ($\pm$ 0.60) & 5.44 ($\pm$ 0.04)&11.39 & 11.40 & 9.87 & 10.64 ($\pm$ 0.90) & 10.93 & 11.72 & 9.00 & 6.72 & 150 \\
\textbf{moonlit}   & 0.00 ($\pm$ 0.00) & 0.00 ($\pm$ 0.00) &0.00 & 0.00 & 0.00 & 0.00 ($\pm$ 0.00) & 0.00 & 0.00 & 0.00 & 0.00 & 1 \\
omniquest  & 6.88 ($\pm$ 0.10) &4.53 ($\pm$ 0.09)& 4.80 & 7.08 & 5.79 & 6.87 ($\pm$ 0.15) & 4.98 & 6.20 & 6.55 & 3.10 & 50 \\
pentari   & 0.00 ($\pm$ 0.00) & 0.00 ($\pm$ 0.00) &0.00 & 0.00 & 0.00 & 0.00 ($\pm$ 0.00) & 0.00 & 0.00 & 0.00 & 0.00 & 70 \\
\textbf{snacktime}  & 19.40 ($\pm$ 0.29) & 0.00 ($\pm$ 0.00) &0.00 & 0.00 & 7.84 & 31.75 ($\pm$ 8.62) & 0.00 & 19.25 & 20.14 & 0.50 & 50 \\
\textbf{sorcerer}  & 6.18 ($\pm$ 1.80) & 5.00 ($\pm$ 0.00) &5.00 & 5.03 & 5.73 & 5.65($\pm$ 1.45) & 11.57 & 5.00 & 5.00 &  5.00 & 400 \\
spellbrkr   & 39.99 ($\pm$ 0.01) & 39.92 ($\pm$ 0.03)&39.94 & 39.97 & 39.86 & 40.00 ($\pm$ 0.00) & 40.00 & 39.96 & 40.00 & 36.20 & 600 \\
spirit   & 1.36 ($\pm$ 0.03) & 0.64 ($\pm$ 0.07)& 1.78 & 1.23 & 1.32 & 1.23 ($\pm$ 0.05) & 1.85 & 1.51 & 1.21 & 0.20 & 250 \\
temple   & 0.00 ($\pm$ 0.00) & 0.00 ($\pm$ 0.00)& 0.00 & 0.00 & 0.00 & 3.52 ($\pm$ 1.99) & 0.00 & 0.00 & 0.00 & 0.00 & 35 \\
zenon   & 0.00 ($\pm$ 0.00) & 0.00 ($\pm$ 0.00)& 0.00 & 0.00 & 0.00 & 0.00 ($\pm$ 0.00) & 0.00 & 0.00 & 0.00 & 0.00 & 20 \\
\textbf{zork1}   & 30.39 ($\pm$ 3.01) & 24.76 ($\pm$ 0.52)&11.30 & 22.75 & 27.44 & 32.17 ($\pm$ 4.39) & 12.70 & 31.36 & 29.10 & 2.40 & 350 \\
\textbf{zork3}   & 0.53 ($\pm$ 0.08) & 0.02 ($\pm$ 0.01)& 0.89 & 0.79 & 0.34 & 0.46 ($\pm$ 0.06) & 0.97 & 0.49 & 0.26 & 0.07 & 7 \\
ztuu  & 3.74 ($\pm$ 0.30) & 0.00 ($\pm$ 0.00)&0.00 & 5.66 & 4.85 & 3.93 ($\pm$ 0.07) & 0.00 & 3.73 & 4.38 & 0.55 & 100 \\
        \end{tabular}
        }
    \caption{Raw scores for variants of \model (GPT-2) on each game. Games in bold are those with ClubFloyd scripts. Note that some scores are only based on one trial. \model (GPT-2), \model (ngram) and \model (w/ Jericho) are based on five trails and the standard deviation is given.}
        \label{tab:variant-pergame}
\end{table*}

\begin{figure*}
\begin{lstlisting}
step 22 
state: [CLS] living room above the trophy case hangs an elvish sword of great antiquity.   [SEP] get sword [SEP] taken.  you are carrying:   a sword   a nasty knife   a rope   a brass lantern   a clove of garlic   a jewel-encrusted egg  living room you are in the living room. there is a doorway to the east, a wooden door with strange gothic lettering to the west, which appears to be nailed shut, a trophy case, and a large oriental rug in the center of the room.   [SEP] 
 gpt2_acts: ['east', 'west', 'north', 'south', 'up', 'down', 'drop sword', 'open trophy case', 'open door', 'southeast', 'wait', 'southwest', 'northwest', 'wear sword', 'northeast', 'out', 'take sword', 'knock on door', 'get statue', 'open trophy', 'get rug', 'close door', 'take rug', 'get all', 'get sword', 'open case', 'take all', 'put sword in case', 'get trophy', 'open gothic'] 
 ngram_acts: ['north', 'east', 'south', 'west', 'open door', 'examine door', 'take all', 'unlock door', 'get all', 'close door', 'drop all', 'put all', 'tie rope', 'examine knife', 'take knife', 'examine case', 'examine sword', 'open case', 'examine rug', 'examine rope', 'examine west', 'take rope', 'take sword', 'examine lantern', 'put knife', 'pull rope', 'take lantern', 'examine egg', 'take rug', 'look under rug'] 
 valid_acts: ['east', 'open egg with lantern', 'throw rope at egg', 'throw egg at knife', 'throw sword at egg', 'throw garlic at egg', 'throw lantern at egg', 'throw knife at egg', 'throw knife at lantern', 'push rug', 'put down all', 'put down rope', 'put down egg', 'put down sword', 'put down garlic', 'put down lantern', 'put down knife', 'take on egg', 'open case', 'turn on lantern'] 
 gold_act: ['push rug'] 
 score: 15


step 23 
state: [CLS] taken.   [SEP] move rug [SEP] with a great effort, the rug is moved to one side of the room, revealing the dusty cover of a closed trap door.  you are carrying:   a sword   a nasty knife   a rope   a brass lantern   a clove of garlic   a jewel-encrusted egg  living room you are in the living room. there is a doorway to the east, a wooden door with strange gothic lettering to the west, which appears to be nailed shut, a trophy case, and a closed trap door at your feet.   [SEP] 
 gpt2_acts: ['east', 'west', 'open door', 'north', 'open trap door', 'south', 'open case', 'down', 'up', 'wait', 'open trap', 'pull rug', 'move cover', 'knock on door', 'southeast', 'push rug', 'search cover', 'open cover', 'out', 'close trap', 'southwest', 'move rug', 'enter trap', 'open gothic', 'drop sword', 'search rug', 'northwest', 'close trap door', 'take rug', 'take all'] 
 ngram_acts: ['north', 'east', 'south', 'west', 'open door', 'examine door', 'take all', 'unlock door', 'get all', 'close door', 'drop all', 'put all', 'tie rope', 'examine knife', 'take knife', 'examine case', 'examine sword', 'open case', 'examine rug', 'examine rope', 'examine west', 'take rope', 'take sword', 'examine lantern', 'put knife', 'pull rope', 'take lantern', 'examine egg', 'take rug', 'look under rug'] 
 valid_acts: ['east', 'open egg with lantern', 'throw rope at egg', 'throw egg at knife', 'throw sword at egg', 'throw garlic at egg', 'throw lantern at egg', 'throw knife at egg', 'throw knife at lantern', 'put down all', 'put down rope', 'put down egg', 'put down sword', 'put down garlic', 'put down lantern', 'put down knife', 'take on egg', 'open trap', 'open case', 'turn on lantern'] 
 gold_act: ['open trap'] 
 score: 15


step 24 
state: [CLS] with a great effort, the rug is moved to one side of the room, revealing the dusty cover of a closed trap door.   [SEP] open trapdoor [SEP] the door reluctantly opens to reveal a rickety staircase descending into darkness.  you are carrying:   a sword   a nasty knife   a rope   a brass lantern   a clove of garlic   a jewel-encrusted egg  living room you are in the living room. there is a doorway to the east, a wooden door with strange gothic lettering to the west, which appears to be nailed shut, a trophy case, and a rug lying beside an open trap door.   [SEP] 
 gpt2_acts: ['east', 'west', 'down', 'up', 'north', 'south', 'open trophy case', 'wait', 'knock on door', 'take rug', 'southeast', 'enter trapdoor', 'out', 'drop sword', 'take rope', 'in', 'southwest', 'northwest', 'get rope', 'open case', 'get rug', 'search rug', 'enter trap', 'climb rope', 'northeast', 'take sword', 'move rug', 'take all', 'put sword in trapdoor', 'close trapdoor'] 
 ngram_acts: ['north', 'east', 'south', 'west', 'open door', 'examine door', 'take all', 'unlock door', 'get all', 'close door', 'drop all', 'put all', 'tie rope', 'examine knife', 'take knife', 'examine case', 'examine sword', 'open case', 'examine rope', 'examine west', 'take rope', 'take sword', 'examine lantern', 'put knife', 'pull rope', 'take lantern', 'examine egg', 'put sword', 'get sword', 'put egg'] 
 valid_acts: ['east', 'open egg with lantern', 'throw rope at egg', 'throw egg at knife', 'throw sword at egg', 'throw garlic at egg', 'throw lantern at egg', 'throw knife at egg', 'throw knife at lantern', 'put down all', 'put down rope', 'put down egg', 'put down sword', 'put down garlic', 'put down lantern', 'put down knife', 'close trap', 'take on egg', 'open case', 'turn on lantern', 'down'] 
 gold_act: ['down'] 
 score: 15


step 25 
state: [CLS] the door reluctantly opens to reveal a rickety staircase descending into darkness.   [SEP] down [SEP] you have moved into a dark place. the trap door crashes shut, and you hear someone barring it.  it is pitch black. you are likely to be eaten by a grue. your sword is glowing with a faint blue glow.  you are carrying:   a sword   a nasty knife   a rope   a brass lantern   a clove of garlic   a jewel-encrusted egg  it is pitch black. you are likely to be eaten by a grue.   [SEP] 
 gpt2_acts: ['down', 'west', 'east', 'north', 'wait', 'south', 'up', 'open door', 'southeast', 'listen', 'southwest', 'out', 'northeast', 'open trap door', 'northwest', 'enter trap', 'drop sword', 'sleep', 'close door', 'knock on door', 'get rope', 'open trap', 'turn off lamp', 'sing', 'stand', 'take rope', 'forward', 'shout', 'pull rope', 'sound'] 
 ngram_acts: ['north', 'east', 'south', 'west', 'open door', 'examine door', 'take all', 'unlock door', 'get all', 'close door', 'drop all', 'put all', 'tie rope', 'examine knife', 'take knife', 'examine case', 'examine sword', 'open case', 'examine rope', 'examine west', 'take rope', 'take sword', 'examine lantern', 'put knife', 'pull rope', 'take lantern', 'examine egg', 'put sword', 'get sword', 'put egg'] 
 valid_acts: ['south', 'north', 'open egg with lantern', 'throw rope at egg', 'throw egg at sword', 'throw garlic at egg', 'throw lantern at egg', 'throw knife at egg', 'throw sword at egg', 'throw sword at lantern', 'put down all', 'put down rope', 'put down egg', 'put down garlic', 'put down lantern', 'put down knife', 'put down sword', 'take on egg', 'turn on lantern', 'east'] 
 gold_act: ['turn on lantern'] 
 score: 40
\end{lstlisting}
\caption{A piece of walkthrough evaluation in Zork1. }
\label{fig:wt_zork1}
\end{figure*}

\newpage

\begin{figure*}
\begin{lstlisting}
State  62235: [CLS]  North of House You are facing the north side of a white house. There is no door here, and all the windows are boarded up. To the north a narrow path winds through the trees. [SEP]  north [SEP]  Forest Path This is a path winding through a dimly lit forest. The path heads north south here. One particularly large tree with some low branches stands at the edge of the path.  You are empty handed.  Forest Path This is a path winding through a dimly lit forest. The path heads north south here. One particularly large tree with some low branches stands at the edge of the path. [SEP]
Actions62235: ['climb tree', 'up', 's', 'n', 'north', 'south', 'east', 'west']
Qvalues62235: [15.38, 15.29, 12.4, 12.34, 11.99, 11.73, 11.13, 10.57]
>> Action62235: up
Reward62235: 0, Score 0, Done False

State  62236: [CLS]  Forest Path This is a path winding through a dimly lit forest. The path heads north south here. One particularly large tree with some low branches stands at the edge of the path. [SEP]  up [SEP]  Up a Tree You are about 10 feet above the ground nestled among some large branches. The nearest branch above you is above your reach. Beside you on the branch is a small bird's nest. In the bird's nest is a large egg encrusted with precious jewels, apparently scavenged by a childless songbird. The egg is covered with fine gold inlay, and ornamented in lapis lazuli and mother of pearl. Unlike most eggs, this one is hinged and closed with a delicate looking clasp. The egg appears extremely fragile.  You are empty handed.  Up a Tree You are about 10 feet above the ground nestled among some large branches. The nearest branch above you is above your reach. Beside you on the branch is a small bird's nest. In the bird's nest is a large egg encrusted with precious jewels, apparently scavenged by a childless songbird. The egg is covered with fine gold inlay, and ornamented in lapis lazuli and mother of pearl. Unlike most eggs, this one is hinged and closed with a delicate looking clasp. The egg appears extremely fragile. [SEP]
Actions62236: ['get egg', 'take egg', 'eat egg', 'down', 'take all', 'get nest', 'take nest']
Qvalues62236: [17.03, 17.02, 16.99, 13.22, 11.75, 11.05, 10.68]
>> Action62236: eat egg
Reward62236: 5, Score 5, Done False

State  62237: [CLS]  Up a Tree You are about 10 feet above the ground nestled among some large branches. The nearest branch above you is above your reach. Beside you on the branch is a small bird's nest. In the bird's nest is a large egg encrusted with precious jewels, apparently scavenged by a childless songbird. The egg is covered with fine gold inlay, and ornamented in lapis lazuli and mother of pearl. Unlike most eggs, this one is hinged and closed with a delicate looking clasp. The egg appears extremely fragile. [SEP]  eat egg [SEP]  (Taken) I don't think that the jewel encrusted egg would agree with you.  You are carrying:   A jewel encrusted egg  Up a Tree You are about 10 feet above the ground nestled among some large branches. The nearest branch above you is above your reach. Beside you on the branch is a small bird's nest. [SEP]
Actions62237: ['down', 'jump', 'put egg in nest', 'drop egg', 'throw egg at nest', 'eat nest', 'get nest', 'take nest']
Qvalues62237: [12.93, 12.93, 11.49, 11.22, 11.1, 9.49, 9.41, 9.36]
>> Action62237: down
Reward62237: 0, Score 5, Done False

State  62238: [CLS]  (Taken) I don't think that the jewel encrusted egg would agree with you. [SEP]  down [SEP]  Forest Path  You are carrying:   A jewel encrusted egg  Forest Path This is a path winding through a dimly lit forest. The path heads north south here. One particularly large tree with some low branches stands at the edge of the path. [SEP]
Actions62238: ['south', 'climb tree', 'drop egg', 'drop all', 'up', 'throw egg at tree', 'east', 'north', 'west']
Qvalues62238: [14.26, 11.78, 11.46, 11.41, 11.03, 11.03, 10.85, 10.72, 10.5]
>> Action62238: south
Reward62238: 0, Score 5, Done False

State  62239: [CLS]  Forest Path [SEP]  south [SEP]  North of House  You are carrying:   A jewel encrusted egg  North of House You are facing the north side of a white house. There is no door here, and all the windows are boarded up. To the north a narrow path winds through the trees. [SEP]
Actions62239: ['east', 'southeast', 'west', 'north', 'southwest', 'drop all', 'drop egg']
Qvalues62239: [15.96, 15.82, 12.94, 12.86, 12.68, 12.32, 12.26]
>> Action62239: southeast
Reward62239: 0, Score 5, Done False

State  62240: [CLS]  North of House [SEP]  southeast [SEP]  Behind House You are behind the white house. A path leads into the forest to the east. In one corner of the house there is a small window which is slightly ajar.  You are carrying:   A jewel encrusted egg  Behind House You are behind the white house. A path leads into the forest to the east. In one corner of the house there is a small window which is slightly ajar. [SEP]
Actions62240: ['drop egg', 'drop all', 'northwest', 'north', 'open window', 'south', 'southwest', 'east']
Qvalues62240: [17.9, 17.89, 14.71, 14.5, 13.59, 13.51, 12.97, 12.8]
>> Action62240: drop all
Reward62240: 0, Score 5, Done False

State  62241: [CLS]  Behind House You are behind the white house. A path leads into the forest to the east. In one corner of the house there is a small window which is slightly ajar. [SEP]  drop all [SEP]  jewel encrusted egg: Dropped.  You are empty handed.  Behind House You are behind the white house. A path leads into the forest to the east. In one corner of the house there is a small window which is slightly ajar. There is a jewel encrusted egg here. [SEP]
Actions62241: ['open window', 'take it', 'get egg', 'get encrusted egg', 'eat egg', 'take egg', 'get it', 'take all', 'get all', 'north', 'northwest', 'south', 'east', 'southwest']
Qvalues62241: [19.91, 16.52, 16.44, 16.4, 16.26, 16.25, 14.9, 14.25, 14.03, 13.86, 13.17, 12.49, 12.45, 12.1]
>> Action62241: open window
Reward62241: 0, Score 5, Done False

State  62242: [CLS]  jewel encrusted egg: Dropped. [SEP]  open window [SEP]  With great effort, you open the window far enough to allow entry.  You are empty handed.  Behind House You are behind the white house. A path leads into the forest to the east. In one corner of the house there is a small window which is open. There is a jewel encrusted egg here. [SEP]
Actions62242: ['west', 'in', 'enter window', 'north', 'northwest', 'east', 'eat egg', 'take egg', 'get egg', 'down', 'get all', 'south', 'southwest', 'up']
Qvalues62242: [22.42, 20.53, 18.46, 16.62, 15.93, 15.17, 14.3, 14.13, 14.1, 13.82, 13.74, 13.68, 13.56, 11.59]
>> Action62242: west
Reward62242: 10, Score 15, Done False

State  62243: [CLS]  With great effort, you open the window far enough to allow entry. [SEP]  west [SEP]  Kitchen You are in the kitchen of the white house. A table seems to have been used recently for the preparation of food. A passage leads to the west and a dark staircase can be seen leading upward. A dark chimney leads down and to the east is a small window which is open. On the table is an elongated brown sack, smelling of hot peppers. A bottle is sitting on the table. The glass bottle contains:   A quantity of water  You are empty handed.  Kitchen You are in the kitchen of the white house. A table seems to have been used recently for the preparation of food. A passage leads to the west and a dark staircase can be seen leading upward. A dark chimney leads down and to the east is a small window which is open. On the table is an elongated brown sack, smelling of hot peppers. A bottle is sitting on the table. The glass bottle contains:   A quantity of water [SEP]
Actions62243: ['open sack', 'eat sack', 'open bottle', 'take sack', 'get sack', 'out', 'enter window', 'east', 'west', 'get bottle', 'take bottle', 'take all', 'get all', 'up']
Qvalues62243: [13.74, 13.68, 12.38, 11.53, 11.4, 11.25, 11.13, 11.06, 10.36, 10.23, 10.15, 9.63, 9.61, 6.54]
>> Action62243: open sack
Reward62243: 0, Score 15, Done False
\end{lstlisting}
\end{figure*}
\newpage
\begin{figure*}
\vspace{-5pt}
\begin{lstlisting}

State  62244: [CLS]  Kitchen You are in the kitchen of the white house. A table seems to have been used recently for the preparation of food. A passage leads to the west and a dark staircase can be seen leading upward. A dark chimney leads down and to the east is a small window which is open. On the table is an elongated brown sack, smelling of hot peppers. A bottle is sitting on the table. The glass bottle contains:   A quantity of water [SEP]  open sack [SEP]  Opening the brown sack reveals a lunch, and a clove of garlic.  You are empty handed.  Kitchen You are in the kitchen of the white house. A table seems to have been used recently for the preparation of food. A passage leads to the west and a dark staircase can be seen leading upward. A dark chimney leads down and to the east is a small window which is open. A bottle is sitting on the table. The glass bottle contains:   A quantity of water There is a brown sack here. The brown sack contains:   A lunch   A clove of garlic [SEP]
Actions62244: ['take sack', 'get sack', 'eat sack', 'take bag', 'take garlic', 'out', 'close sack', 'get garlic', 'take lunch', 'take clove', 'eat garlic', 'east', 'west', 'take bottle', 'get all', 'take all', 'up']
Qvalues62244: [15.32, 15.25, 15.23, 14.95, 12.16, 12.12, 11.9, 11.89, 11.84, 11.7, 11.66, 11.65, 11.18, 10.95, 10.44, 10.39, 9.46]
>> Action62244: get sack
Reward62244: 0, Score 15, Done False

State  62245: [CLS]  Opening the brown sack reveals a lunch, and a clove of garlic. [SEP]  get sack [SEP]  Taken.  You are carrying:   A brown sack   The brown sack contains:     A lunch     A clove of garlic  Kitchen You are in the kitchen of the white house. A table seems to have been used recently for the preparation of food. A passage leads to the west and a dark staircase can be seen leading upward. A dark chimney leads down and to the east is a small window which is open. A bottle is sitting on the table. The glass bottle contains:   A quantity of water [SEP]
Actions62245: ['get clove', 'take clove', 'get garlic', 'take garlic', 'west', 'put garlic on table', 'drop garlic', 'open bottle', 'put sack on table', 'get food', 'drop all', 'drop sack', 'out', 'east', 'get bottles', 'take bottle', 'get bottle', 'take glass', 'get all', 'up']
Qvalues62245: [16.25, 16.19, 16.01, 15.94, 14.53, 14.18, 14.04, 13.83, 12.39, 11.77, 11.62, 11.56, 11.05, 10.86, 10.26, 10.21, 10.08, 9.4, 9.3, 7.84]
>> Action62245: take clove
Reward62245: 0, Score 15, Done False

State  62246: [CLS]  Taken. [SEP]  take clove [SEP]  Taken.  You are carrying:   A clove of garlic   A brown sack   The brown sack contains:     A lunch  Kitchen You are in the kitchen of the white house. A table seems to have been used recently for the preparation of food. A passage leads to the west and a dark staircase can be seen leading upward. A dark chimney leads down and to the east is a small window which is open. A bottle is sitting on the table. The glass bottle contains:   A quantity of water [SEP]
Actions62246: ['west', 'open bottle', 'drop clove', 'put garlic on table', 'drop garlic', 'put clove on table', 'put clove in sack', 'put garlic in sack', 'east', 'out', 'take bottle', 'get bottle', 'get all', 'up', 'get glass', 'take glass']
Qvalues62246: [17.59, 16.09, 14.8, 14.69, 14.66, 14.47, 13.37, 13.26, 12.45, 12.12, 11.82, 11.51, 10.67, 10.48, 10.48, 10.38]
>> Action62246: west
Reward62246: 0, Score 15, Done False

State  62247: [CLS]  Taken. [SEP]  west [SEP]  Living Room You are in the living room. There is a doorway to the east, a wooden door with strange gothic lettering to the west, which appears to be nailed shut, a trophy case, and a large oriental rug in the center of the room. Above the trophy case hangs an elvish sword of great antiquity. A battery powered brass lantern is on the trophy case.  You are carrying:   A clove of garlic   A brown sack   The brown sack contains:     A lunch  Living Room You are in the living room. There is a doorway to the east, a wooden door with strange gothic lettering to the west, which appears to be nailed shut, a trophy case, and a large oriental rug in the center of the room. Above the trophy case hangs an elvish sword of great antiquity. A battery powered brass lantern is on the trophy case. [SEP]
Actions62247: ['light lantern', 'get lantern', 'take lantern', 'turn off lantern', 'open case', 'open trophy case', 'turn on lantern', 'east', 'take all', 'get sword', 'get all', 'take sword', 'drop all']
Qvalues62247: [20.04, 19.1, 18.96, 18.95, 18.35, 18.35, 17.76, 15.08, 14.41, 14.23, 14.12, 13.97, 12.25]
>> Action62247: light lantern
Reward62247: 0, Score 15, Done False

State  62248: [CLS]  Living Room You are in the living room. There is a doorway to the east, a wooden door with strange gothic lettering to the west, which appears to be nailed shut, a trophy case, and a large oriental rug in the center of the room. Above the trophy case hangs an elvish sword of great antiquity. A battery powered brass lantern is on the trophy case. [SEP]  light lantern [SEP]  (Taken) The brass lantern is now on.  You are carrying:   A brass lantern (providing light)   A clove of garlic   A brown sack   The brown sack contains:     A lunch  Living Room You are in the living room. There is a doorway to the east, a wooden door with strange gothic lettering to the west, which appears to be nailed shut, a trophy case, and a large oriental rug in the center of the room. Above the trophy case hangs an elvish sword of great antiquity. [SEP]
Actions62248: ['take sword', 'get sword', 'turn off lantern', 'drop sack', 'drop all', 'drop lantern', 'east']
Qvalues62248: [22.64, 22.46, 19.89, 17.26, 15.66, 15.57, 14.99]
>> Action62248: get sword
Reward62248: 0, Score 15, Done False

State  62249: [CLS]  (Taken) The brass lantern is now on. [SEP]  get sword [SEP]  Taken.  You are carrying:   A sword   A brass lantern (providing light)   A clove of garlic   A brown sack   The brown sack contains:     A lunch  Living Room You are in the living room. There is a doorway to the east, a wooden door with strange gothic lettering to the west, which appears to be nailed shut, a trophy case, and a large oriental rug in the center of the room. [SEP]
Actions62249: ['move rug', 'drop sword', 'turn off lantern', 'east', 'put lantern in case', 'up']
Qvalues62249: [25.33, 15.03, 13.93, 11.99, 11.17, 10.45]
>> Action62249: move rug
Reward62249: 0, Score 15, Done False

State  62250: [CLS]  Taken. [SEP]  move rug [SEP]  With a great effort, the rug is moved to one side of the room, revealing the dusty cover of a closed trap door.  You are carrying:   A sword   A brass lantern (providing light)   A clove of garlic   A brown sack   The brown sack contains:     A lunch  Living Room You are in the living room. There is a doorway to the east, a wooden door with strange gothic lettering to the west, which appears to be nailed shut, a trophy case, and a closed trap door at your feet. [SEP]
Actions62250: ['open trap door', 'open trapdoor', 'open cover', 'drop lantern', 'east']
Qvalues62250: [27.97, 27.94, 27.91, 21.05, 20.41]
>> Action62250: open cover
Reward62250: 0, Score 15, Done False

State  62251: [CLS]  With a great effort, the rug is moved to one side of the room, revealing the dusty cover of a closed trap door. [SEP]  open cover [SEP]  The door reluctantly opens to reveal a rickety staircase descending into darkness.  You are carrying:   A sword   A brass lantern (providing light)   A clove of garlic   A brown sack   The brown sack contains:     A lunch  Living Room You are in the living room. There is a doorway to the east, a wooden door with strange gothic lettering to the west, which appears to be nailed shut, a trophy case, and a rug lying beside an open trap door. [SEP]
Actions62251: ['down', 'close trap door', 'east', 'drop sack', 'drop lantern', 'put sack in case']
Qvalues62251: [30.88, 25.45, 22.21, 21.96, 20.7, 20.62]
>> Action62251: down
Reward62251: 25, Score 40, Done False
\end{lstlisting}
\caption{Last episode of game trajectory of DRRN + \model (GPT-2) on Zork 1, from start until 40 is scored.}
\label{fig:end}
\end{figure*}
\newpage
\begin{figure*}
\begin{lstlisting}
Example 1: "You see the monster stumble from its cave. You are carrying a bow and three arrows"

CALM (GPT-2) Top 10 Generated Actions: ['south',
 'hit monster with bow',
 'up',
 'shoot monster with bow',
 'down',
 'east',
 'west',
 'north',
 'kill monster',
 'shoot monster']

Example 2: "Tom looked concerned. The panel of levers and dials clearly was confusing him"

CALM (GPT-2) Top 10 Generated Actions: ['south',
 'pull lever',
 'talk to tom',
 'open panel',
 'east',
 'west',
 'turn dials',
 'north',
 'push button',
 'pull levers']

Example 3:"Your body feels cold as you plunge into the river"

CALM (GPT-2) Top 10 Generated Actions: ['south',
 'wait',
 'up',
 'enter river',
 'down',
 'east',
 'west',
 'swim',
 'drink water',
 'north']
\end{lstlisting}
\caption{Some handpicked example observations and top 10 action predictions for CALM (GPT-2). The top non-directional actions demonstrate some understanding of the objects present in the observations, and some commonsense actions involving those objects.}
\end{figure*}

\end{document}